# Review of Fruit Tree Image Segmentation

Il-Seok Oh *

Department of Computer Science and Artificial Intelligence/CAIIT, Jeonbuk National University, South Korea

isoh@jbnu.ac.kr

Fruit tree image segmentation is an essential problem in automating a variety of agricultural tasks such as phenotyping, harvesting, spraying, and pruning. Many research papers have proposed a diverse spectrum of solutions suitable to specific tasks and environments. The review scope of this paper is confined to the front views of fruit trees and based on 158 relevant papers collected using a newly designed crawling review method. These papers are systematically reviewed based on a taxonomy that sequentially considers the method, image, task, and fruit. This taxonomy will assist readers to intuitively grasp the big picture of these research activities. Our review reveals that the most noticeable deficiency of the previous studies was the lack of a versatile dataset and segmentation model that could be applied to a variety of tasks and environments. Six important future research tasks are suggested, with the expectation that these will pave the way to building a versatile tree segmentation module.

## 1 INTRODUCTION

To maximize the production of fruit trees in orchards, farmers and horticulturists perform various tasks such as phenotyping, health and growth monitoring, spraying, pruning, and yield estimation. In the past, these tasks were performed manually. However, the fatigue of the human laborers often led to imprecise and error-prone results. In addition, the continual increase in labor costs has resulted in high market prices for fruits. The automation of these tasks has become very important and garnered much attention from agronomists and computer scientists [Kamilaris 2018, Thakur 2023].

Computer-vision researchers have continually attempted to develop automatic systems for these agricultural tasks [Dhanya 2022, Luo 2023]. However, the large variations in the tree arrangements in orchards, tree shapes, weather, and image acquisition devices have made traditional computer vision algorithms far from practical. Recently, owing to the advent of deep learning technology, practical systems are possible, and commercial systems are available. Deep learning has transformed the fundamentals of computer vision from manual design through human reasoning to machine learning-based design through neural network optimization [Zhang 2023a].

---

Il-Seok Oh, Department of Computer Science and Artificial Intelligence/Center for Advanced Image and Information Technology, Jeonbuk National University, South Korea; e-mail: isoh@jbnu.ac.kr

Obtaining relevant information about individual trees through detailed observations of their trunks, branches, leaves, and fruit is essential to successfully accomplish the necessary tasks. Therefore, an important preliminary problem in automating these tasks is tree image segmentation [Chehreh 2023]. Segmenting the regions containing individual trees is required. Often a finer segmentation of the tree region into the trunk, branches, leaves, and fruit must also be performed. The tree segmentation problem is challenging because of several factors. First, the acquired images have large variations. Many factors influence these variations, including the fruit type, geography, weather, and farming techniques. Second, occlusion by various facilities such as trellises and poles commonly occurs. The self occlusion produced by leaves and branches makes the segmentation problem difficult. Third, because fruit trees are placed in a row and separated by equal distances, their branches are commonly intertwined. The boundaries between adjacent trees are ambiguous.

Recently, a survey paper on tree image segmentation was published [Chehreh 2023]. Although the paper considered images from orchards and a forest, it gave more attention to the top-view images acquired by unmanned aerial vehicles (UAVs). Therefore, the application tasks made possible by the segmentation results were oriented toward the digital forestry domain. In contrast, agricultural tasks such as harvesting, spraying, and pruning require the segmentation of the front-view images of trees. To the best of our knowledge, there has been no survey or review paper dealing with fruit tree segmentation in the agricultural domain. This fact motivated our study.

In this paper, we present a review of fruit tree image segmentation algorithms, involving both traditional rule-based (RB) and modern deep learning (DL) approaches. The scope of this review is confined to segmenting front-view images of fruit trees. It excludes the segmentation of forest trees, which belongs to the digital forestry domain. Using a newly designed crawling review method, 158 relevant papers were collected. Our review was performed by grouping the papers according to several criteria. The hierarchy of the taxonomy had the following order: the technical approach (RB vs. DL), image type (RGB, RGB-D, point cloud, others), task (harvesting, phenotyping, spraying, pruning, yield estimation, etc.), fruit type (apple, grape, citrus, pear, peach, litchi, guava, cherry, etc.), and publication year (from old to recent).

Section 2 outlines our review, focusing on the review scope, method, and statistics. Section 3 reviews the papers that reported results based on an RB approach. Section 4 reviews the papers that reported results based on DL. Section 5 presents a discussion and future work. Section 6 concludes the paper.

## 2 REVIEW SCOPE, METHOD, AND STATISTICS

Because tree segmentation is a broad research topic, we focus on the agricultural domain. Section 2.1 describes our review scope in detail. Section 2.2 explains how we collected and filtered the related papers, along with the taxonomy used. Section 2.3 gives some statistics about the papers.

### 2.1 Review scope

We can divide tree segmentation into two domains: fruit trees in an agricultural domain and forest trees in a digital forestry domain. Fruit trees are usually planted in orchards in rows. The typical tasks supported by tree segmentation in this domain are related to precision farming and include phenotyping, harvesting, spraying, pruning, yield estimation, and robot navigation. These tasks require the segmenting of individual trees and often the finer segmentation of a tree region into parts such as the trunk, branches, leaves, and fruit. The digital forestry domain treats a larger area where the trees are randomly or regularly planted. It requires a coarse segmentation of trees covering a large forest. Typical tasks include species and habitat identification, population estimation, spraying, health and growth monitoring, and logging planning.

Two different methods are used to capture tree images. In the first, a person or robot captures images in front of target trees, resulting in front-view images. In the second, top-view images are acquired using unmanned aerial vehicles (UAVs) or drones carrying image sensors. Front-view images are the major type utilized in the agricultural domain, while top-view images are primarily used in the digital forestry domain. Figure 1 illustrates examples of front- and top-view tree images.

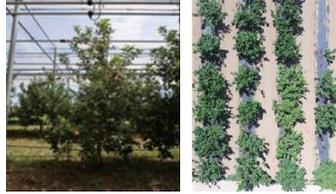

Figure 1: Tree images: front view of apple trees on left [La 2023] and top view of cherry trees on right [Cheng 2020].

Our review is confined to the front-view images of fruit trees. Chronologically, our review covers a long period extending from 1990 to 2023. Methodologically, it includes both RB and DL approaches. The RB algorithms include thresholding, clustering, region growing, machine learning, fitting, and graph-based methods. Papers reporting the results of studies that adopted DL began to appear in the late 2010s. These DL algorithms used convolutional neural network (CNN) models such as mask R-CNN and YOLACT, and transformer models such as DETR and Swin. Sections A.1.2 and A.1.3 briefly explain these algorithms.

We exclude papers that deal only with the segmentation of the fruit, which is regarded as a separate problem. For these fruit-only segmentation algorithms, we refer the readers to another survey paper [Xiao 2023]. Papers that discuss the segmenting of the trunk only, branches only, or flowers only are included. Moreover, papers that discuss the segmenting of the fruit along with the fruit-bearing stems are included. Figure 2 demonstrates several levels of tree segmentation, including whole tree, branch, branch classification, fruit with stem, fruit with picking point, and flower segmentation.

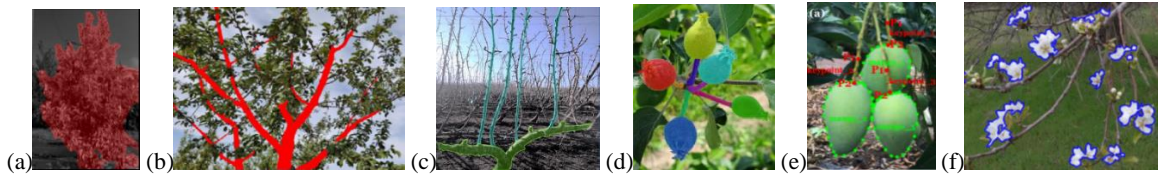

Figure 2: Various levels of tree segmentation: (a) whole tree [La 2023], (b) branch [Wan 2023], (c) branch classification [Borrenpohl 2023], (d) fruit and stem [Hussain 2023], (e) fruit and picking point [Zheng 2021], and flower [Dias 2018a]).

The explanation of the papers in this review focuses on the methodology rather than the performance or effect. Because of the lack of a standard dataset, an objective performance comparison between different methods is currently of little importance.

## 2.2 Literature search and taxonomy

*2.2.1 Search.*

A *systematic review* is the standard method used by most review papers. In a systematic review, the authors collect a preliminary set of relevant papers by searching databases such as the Web of Science (WoS) or Google scholar using a well-designed query [Snyder 2019]. This preliminary set is then filtered by the authors to select the final set of papers, which is the reading and review target. For example, in one review paper [Chehreh 2023] on tree segmentation, the authors searched Google scholar for papers confined to the period of 2013–2023 using the query keywords "UAV," "tree," "segmentation," and "classification." They collected 979 peer-reviewed papers as a preliminary set and filtered them to obtain 144 papers as their final set.

Because of the high variability of fruit tree segmentation, as demonstrated in Figure 2, we assumed that the review required another search method for relevant papers. The lack of a standard dataset and insufficiency of review or survey papers on tree segmentation support this argument. We proposed a novel review method, called a *crawling review*. The search process of a crawling review is similar to that of web crawling. Just as web crawling starts with a queue containing a seed URL, a crawling review starts with a queue containing a seed paper or set of seed papers. The relevant papers cited in the seed paper are then pushed into the queue. Papers are removed from the queue and examined one at a time, and this process of adding papers to the queue and removing papers from it continues until the queue is empty. Figure 3(a) explains this process formally. Figure 3(b) shows the supplementary phase, which collects additional articles from the most recent issues of the most cited journals. This supplementary phase is optional.

In our review, [Chehreh 2023] was the seed paper used to initiate the queue, $Q$. We utilized the supplementary phase, where the $J$ queue was initiated with three journals: *Computers and Electronics in Agriculture*, *Biosystems Engineering*, and the *Journal of Field Robotics*. The given period was set to January 2020–December 2023.

The quality of the conventional review method using a well-designed query is heavily dependent on the query design and search engine. We believe that this novel crawling review will result in a better quality because it is closer to an exhaustive search. However, our method is much more laborious and takes longer because the search process is done manually.

*2.2.2 Taxonomy.*

The papers collected by the process illustrated in Figure 3 are classified according to taxonomy in Figure 4. The classification hierarchy has the following order: the tree type (fruit or forest), view (front or top), method, image type, and task. As mentioned in the previous section, the forest tree and top-view images were excluded from our review. The successive criterion not shown in Figure 4 was the fruit species such as apple, grape, citrus, peach, and guava. The fruit species are listed in Table 1 and Table 2, where the relevant papers are listed.

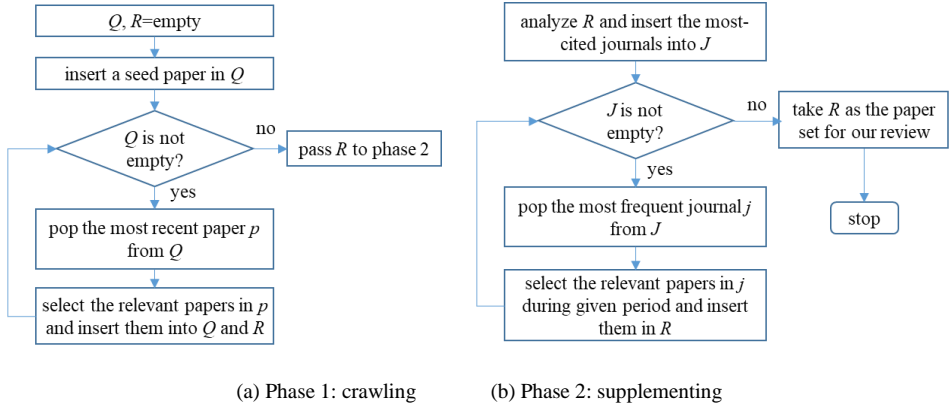

(a) Phase 1: crawling  (b) Phase 2: supplementing

Figure 3: Crawling review with supplementary phase.

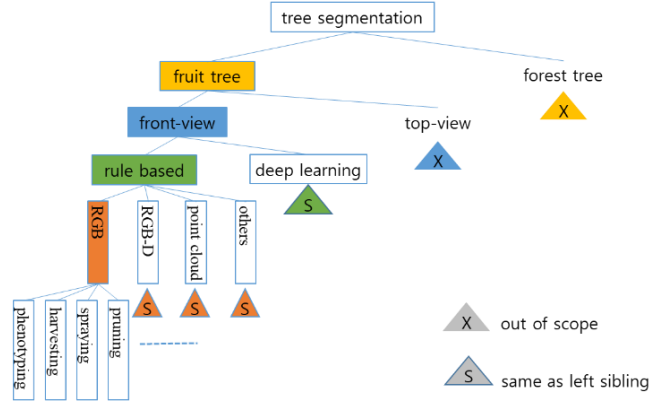

Figure 4: Taxonomy of literature and review scope.

## 2.3 Statistics

Figure 5(a) shows a rapid increase in the number of papers published per year. The RB papers reached a peak in approximately 2015–2018 and decreased afterward. The DL papers began to appear in 2018 and have maintained an increasing trend. This trend demonstrates that the methodological paradigm is fundamentally changing from RB to DL. It also shows that tree segmentation is garnering increasing attention as a primary building block for implementing various agricultural automation tasks. Figure 5(b) shows the statistics of the various data types. In the past, when studies relied on RB methods, similar high ratios of RGB and point cloud methods were used. It seems that the RGB method was adopted because of its low cost, while the point cloud was adopted because of its robustness against illumination. In the DL approach, because robustness can be obtained using RGB images, these have become dominant. The high cost of obtaining a point cloud and multi-spectral images has greatly reduced their use.

Figure 5(c) illustrates the number of papers per agricultural task supported by tree segmentation. Phenotyping and harvesting are the two most popular tasks. One noticeable fact is that in the DL era, harvesting has the highest frequency. This is because DL makes practical and commercial systems possible, and the most urgent task is harvesting because of the high cost of manual harvesting [Droukas 2023]. Figure 5(d) shows the statistics according to the fruit type. Apples and grapes are two most popular fruits. Some of the fruits not shown in the figure are avocado, apricot, Chinese hickory, yucca, pomegranate, longan, and passion fruit.

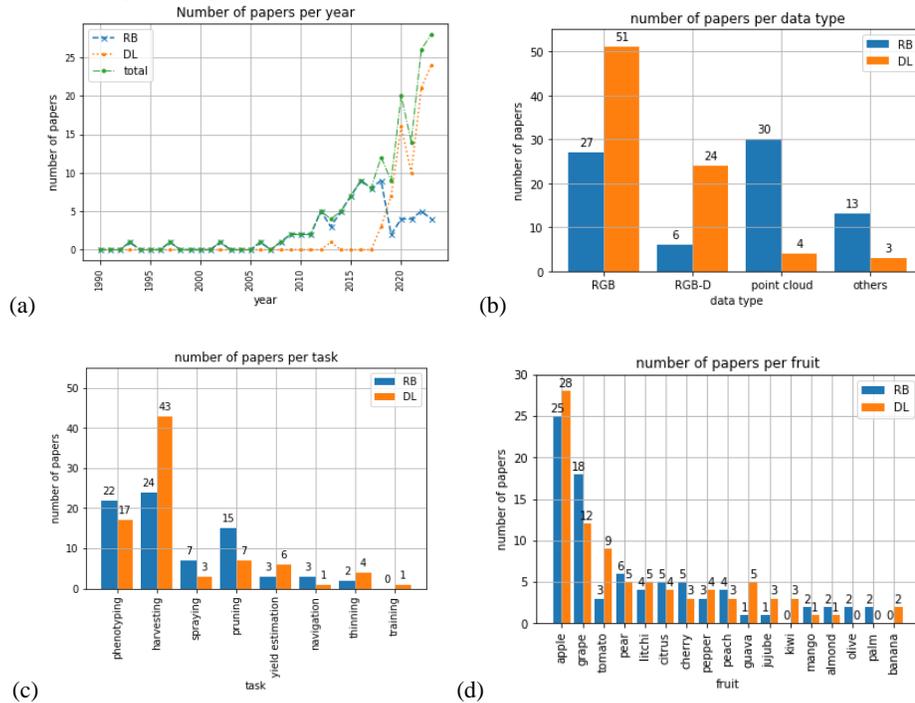

Figure 5: Literature statistics.

## 3 RULE-BASED ALGORITHMS FOR TREE IMAGE SEGMENTATION

Table 1 summarizes the 76 papers that discuss the use of RB segmentation algorithms. It follows the taxonomy hierarchy shown in Figures 4 and 6. The rows are ordered in accordance with the hierarchy level of the image type. In each row, the papers are listed in terms of agricultural tasks, in the order shown in Figure 6(a). For each task, the papers are further listed in order of the fruit type, as shown in Figure 6(b). Each paper shows the fruit type in parentheses. Some papers consider multiple fruit types. The section structure is strictly in accordance with Table 1. The papers discussing the same fruit type are placed in the same paragraph. When a research group published a series of paper, the papers are reviewed together to make it more convenient to track the research progress.

(a) phenotyping→harvesting→spraying→pruning→yield estimation→navigation→thinning→training

(b)apple→grape→tomato→pear→litchi→citrus→cherry→pepper→peach→guava→jujube→kiwi→mango→almond→olive→palm→banana

Figure 6. Order of review: (a) in terms of task and (b) in terms of fruit.

Table 1: Summary of RB tree image segmentation papers

| Image Type | Agricultural Tasks and Relevant Papers |
| --- | --- |
| RGB | **Phenotyping**: [Tabb 2017](apple), [Tabb 2018](apple), [Svensson 2002](grape)<br>**Harvesting**: [Ji 2016a](apple), [Ji 2016b](apple), [Silwal 2017](apple), [Xiang 2018](tomato), [Deng 2011](litchi), [Zhuang 2019](litchi), [Xiong 2018a](litchi), [Xiong 2018b](litchi), [Pla 1993](citrus), [Qiang 2014](citrus), [Liu 2018](citrus), [Amatya 2016](cherry), [Amatya 2017](cherry), [Mohammadi 2023](palm), [He 2012](Chinese hickory), [Wu 2014](Chinese hickory)<br>**Spraying**: [Hocevar 2010](apple), [Berenstein 2010](grape), [Asaei 2019](olive)<br>**Pruning**: [McFarlane 1997](grape), [Gao 2006](grape)<br>**Yield estimation**: [Roy 2018](apple)<br>**Navigation**: [Juman 2016](palm)<br>**Thinning**: [Zhang 2022a](apple) |
| RGB-D | **Phenotyping**: [Xue 2018](pear)<br>**Harvesting**: [Lin 2020](guava, pepper, eggplant)<br>**Spraying**: [Xiao 2017](grape, peach, apricot), [Gao 2018](grape), [Gao 2020](peach)<br>**Navigation**: [Gimenez 2022](pear) |
| Point cloud | **Phenotyping**: [Polo 2009](apple, pear, grape), [Rosell 2009](apple, pear, grape, citrus), [Mendes 2014](apple, pear, grape), [Das 2015](apple, grape), [Peng 2016](apple), [Zhang 2020d](apple), [Wahabzada 2015](grape), [Scholer 2015](grape), [Mack 2017](grape), [Zhu 2023a](tomato), [Zhu 2023b](tomato), [Cheein 2015](pear), [Nielsen 2012a](peach), [Underwood 2015](almond), [Bargoti 2015](apple), [Underwood 2016](almond), [Westling 2021a](mango, avocado)<br>**Harvesting**: [Yongting 2017](apple)<br>**Spraying**: [Mahmud 2021](apple)<br>**Pruning**: [Karkee 2014](apple), [Elfiky 2015](apple), [Zeng 2020](apple), [You 2022a](cherry), [You 2022b](cherry), [Li 2023a](jujube), [Westling 2021b](mango, avocado)<br>**Yield estimation**: [Zine-El-Abidine 2021](apple), [Dey 2012](grape)<br>**Navigation**: [Zhang 2013](NA)<br>**Thinning**: [Nielsen 2012b](peach) |
| Others | **Phenotyping**: [Chene 2012](apple, yucca)<br>**Harvesting**: [Jin 2022](grape), [Qiang 2011](citrus), [Tanigaki 2008](cherry), [Bac 2013](pepper), [Bac 2014](pepper), [Colmenero-Martinez 2018](olive)<br>**Pruning**: [Chattopadhyay 2016](apple), [Akbar 2016](apple), [Medeiros 2017](apple), [Botterill 2013](grape), [Botterill 2017](grape), [Luo 2016](grape) |

### 3.1 RGB

Many studies used RGB images because they are the most popular and cheapest. They used various RB segmentation algorithms, as explained in Section A.1.2.

*3.1.1 Phenotyping.*

Tabb et al. measured tree traits such as the structure, diameter, length, and angle of the branches of apple trees for the phenotyping and pruning tasks [Tabb 2017]. They captured tree images by placing a blue curtain behind the tree to ease the processing of the vision algorithm. A six-stage procedure was applied, which included calibration, segmentation, reconstruction, skeletonization, graph representation, and feature of interest extraction. The same research group improved their work by adopting super-pixel and GMM methods [Tabb 2018].

Svensson et al. proposed a leaf and trunk region segmentation algorithm for the tasks of shoot counting and canopy assessment of grapevines [Svensson 2002]. They placed curtains of various colors, including white, magenta, and blue, and evaluated the effectiveness of those colors. First, thresholding was applied to extract the canopies. Vertical edge detection was used to remove the leaf regions, and trunk regions were detected using the maximum point of the Radon transform.

*3.1.2 Harvesting.*

With the aim of building an apple picking robot, Ji et al. converted the color space from RGB to XYZI1I2I3 [Ji 2016a]. By analyzing the proper threshold ranges for the fruit, branches, leaves, and sky, an image was segmented into different regions. The overall performance was improved by applying the contrast limited adaptive histogram equalization (CLAHE) technique to enhance the contrast. The same research group proposed another method that used mean shift clustering to identify the fruit, leaves, and branches [Ji 2016b]. For apple picking, Silwal et al. captured apple tree images by placing a black curtain behind a tree [Silwal 2017]. To alleviate the illumination variation, five images were combined using the exposure fusion technique. The distance to the target fruit was also identified using a depth camera.

Xiang et al. proposed an algorithm for segmenting tomato plant images [Xiang 2018]. Their method used images captured at night to reduce the illumination variation during the daytime. It first converted a color image into a grayscale image using a green and red aberration technique. The grayscale image was then segmented using a pulse-coupled neural network (PCNN). The paper effectively described the advantage of night vision.

Deng et al. proposed an algorithm that segmented a litchi tree image into fruit-bearing branches [Deng 2011]. They obtained string-like litchi regions by thresholding the Cr channel of the YCbCr color space. The litchi fruits were identified by applying k-means clustering to the string-like litchi regions. The string-like stems were identified using image subtraction and a morphological operation. Zhuang et al. proposed an algorithm for picking litchi [Zhuang 2019]. An iterative retinex algorithm was applied to reduce the illumination effect while keeping the chromatic information. Ripe litchi fruit regions were segmented by applying Otsu thresholding to RG chromatic mapping images. The region was further segmented by analyzing the hue level distributions with a box plot. Xiong et al. proposed a litchi tree segmentation algorithm and picking point determination method [Xiong 2018a]. It used images captured at night to alleviate the illumination effect. In the YIQ color space, fuzzy clustering method was applied to remove the background. Then, using Otsu thresholding, the stem and fruit regions were segmented. Finally, the robot picking points were identified using the Harris corner. The same research group extended their work by using binocular stereo to measure the distance to the target fruit [Xiong 2018b].

Pla et al. proposed a citrus tree segmentation method that identified the fruit, leaf, and sky regions [Pla 1993]. To diminish the illumination and highlight effects, they constructed a 2D directional space using a dichromatic reflection model and applied a clustering algorithm in this space. Qiang et al. proposed a fruit and branch segmentation algorithm for citrus tree images [Qiang 2014]. Representing a pixel with an (r,g,b) feature vector, a multi-class SVM was trained to classify the pixels into four classes: fruit, branches, leaves, and background. The segmentation result was used for fruit detection and path planning for robot harvesting. Liu et al. presented a segmentation algorithm for harvesting citrus trees [Liu 2018]. In the training stage, after converting an RGB image to the Y'CbCr color space, multi-elliptical boundary models were constructed in the Cr-Cb space for each fruit and stem region. In the testing stage, the fruit and stem regions were identified in novel tree images using the model.

With the aim of selecting the shaking point for a shake-and-catch machine, Amatya et al. proposed a segmentation algorithm for cherry tree images [Amatya 2016]. To reduce the illumination effect, the images were captured at night using white LED illumination. Representing a pixel with an (r,g,b) feature vector, a Bayesian model was used to classify the pixels into fruit, branch, leaf, and background regions. After the initial segmentation, the pixels in the branch region were fitted to a curve. The detected curves were merged into a larger and more reliable segment that was used to select the shaking point. The same research group extended their work by using 3D information [Amatya 2017].

Mohammadi et al. proposed a stem and lead area segmentation method for cutting and picking the date bunches of palm trees [Mohammadi 2023]. Two smartphones were used to capture RGB images. The image from the first camera was used for stem segmentation by thresholding each of the RGB channels and combining them. The image from the second camera was used to identify the leaf area. These two sources of information were used to control the actuator's speed and motion to reach the cutting point.

With the aim of building a shake-and-catch harvesting machine, He et al. proposed a 3D branch reconstruction method for Chinese hickory tree images [He 2012]. They used two images taken at different viewpoints. From each image, tree branches were roughly estimated using the snake segmentation algorithm. The branch regions from two images were matched to reconstruct a 3D skeleton of tree branches. The same research group extended their work by presenting an improved segmentation method [Wu 2014]. A method to determine the optimal vibration frequency for the trunk shaker was also presented.

*3.1.3 Spraying.*

Hocevar et al. proposed an approach for spraying apple trees [Hocevar 2010]. The green area was extracted as the tree crown by thresholding each channel of an HSI image, combining the results, and applying the erosion operation. Experiments done with water sensitive papers showed a pesticide reduction of 23%.

Berenstein et al. presented a grapevine segmentation algorithm for an automatic spraying task [Berenstein 2010]. They presented algorithms for foliage and grape cluster segmentation. The foliage segmentation was accomplished by analyzing the green channel because the foliage area was green. Edge distribution was used to segment the grape clusters. Experiments showed a 30% reduction in the pesticide used.

Asaei et al. proposed a segmentation method for olive tree images [Asaei 2019]. Thresholding the green channel made it possible to identify the pixels that belonged to the tree region. When the ratio of green pixels exceeded a preset threshold, the nozzle was opened. Using this simple technique, they obtained a 54% pesticide reduction.

*3.1.4 Pruning.*

McFarlane et al. restricted the domain to trained grapevines that had two long branches connected to a trunk [McFarlane 1997]. The images were captured during winter when the vines were dormant. To facilitate the trunk detection, the image was taken with the tree trunk at a specific location in the image. The wire was identified using thresholding and the Hough transform. The trunk was identified by analyzing a vertical histogram of the thresholded binary image. The branches were identified by noise removal, skeletonization, and line merging operations. Gao et al. used a white curtain behind a grapevine to easily segment branch regions by thresholding [Gao 2006]. By using a rule set, the canes were located first and then the nodes were found. Using that information, branch cutting points were identified.

*3.1.5 Yield Estimation.*

Roy et al. proposed a 3D reconstruction method for rows of trees in an apple orchard for yield estimation or phenotyping [Roy 2018]. Each of the frontside and backside images was processed to obtain point clouds. The main problem was to combine a pair of point clouds to reconstruct a 3D tree row model. Based on the constraints that the occlusion boundary from two images were the same and trunks were at similar heights, the pair was optimally aligned under an objective function using various techniques such as principal component analysis (PCA), RANSAC, bundle adjustment, and GMM.

*3.1.6 Navigation.*

With the aim of building an autonomous robot navigation system, Juman et al. proposed a method based on the Viola-Jones algorithm to segment the tree trunk in a palm tree image [Juman 2016]. The Viola-Jones algorithm was first proposed for the purpose of face detection using the AdaBoost classification model. The paper modified the algorithm to improve the segmentation accuracy using depth information.

*3.1.7 Thinning.*

With the aim of flower thinning, Zhang et al. proposed a 3D tree reconstruction method for apple tree rows using three RGB images [Zhang 2022a]. Left, right, and top images were acquired using a UAV. Based on PCA, three images were aligned to reconstruct a 3D tree model. The flower regions were segmented using a thresholding technique and mapped to the 3D tree model. Finally, the thinning plan was made based on the intensity of tree flowering.

## 3.2 RGB-D

Because of the availability and low cost of RGB-D cameras after the release of Kinect in 2010, research based on RGB-D images has rapidly increased. A common strategy of methods based on RGB-D images is to enhance the accuracy of the segmentation using additional depth information. The studies differed in how the depth information was used.

*3.2.1 Phenotyping.*

Xue et al. proposed a trunk detection method for pear trees [Xue 2018]. They acquired RGB images and separately obtained depth images using a SICK LMS 92 laser scanner. An approach to align the RGB and depth images was described in detail. The color image was used to segment the trunk of a tree. The trunk width was measured using the depth image. These two sources of information were fused using the Dempster-Shafer theory.

*3.2.2 Harvesting.*

With the aim of building a picking robot, Lin et al. proposed a segmentation algorithm for guava, pepper, and eggplant images [Lin 2020]. Kinect v2 was used to acquire RGB-D images. First, the RGB image was segmented using a probabilistic segmentation algorithm that relied on prior probabilities for the foreground (fruit) and background (leaves, branches, soil, and sky). The filtered image obtained by multiplying the segmented binary image and depth image contained both foreground and background regions. To keep only the foreground, region growing was applied and a 3D model was recovered using the M-estimator sample consensus (MSAC) method. Finally, SVM was used to classify the foreground and background regions.

*3.2.3 Spraying.*

Xiao et al. proposed a leaf wall segmentation method for spraying devices [Xiao 2017]. Vineyard, peach, and apricot images acquired with Kinect v1 were used. Assuming that the leaf color was mainly green, the green channel was thresholded to obtain a pre-segmented binary map. An improved segmentation map was obtained by combining the binary and depth maps. The spraying control module calculated the distance between the obtained leaf wall and spraying device to control the amount of pesticide. The same research group improved the system by acquiring front and side images of a vineyard using two Kinect cameras [Gao 2018]. They combined the two segmentation results to improve the accuracy.

The same research group proposed another improved segmentation method that combined the color and depth segmentation maps and applied their method to a peach tree image [Gao 2020]. The color image was segmented by thresholding the green channel. The depth channel was segmented using the k-means clustering algorithm. Using a simple rule, two binary maps were combined to obtain the final segmentation of the leaf wall area.

*3.2.4 Navigation.*

With the aim of facilitating autonomous navigation along tree rows, Gimenez et al. proposed a trunk segmentation method for RGB-D video captured in a pear orchard [Gimenez 2022]. The trunk region was obtained by using a series of processes to discard non-trunk pixels. Assuming that the trees were planted in a regular pattern and the robot had a linear velocity, a simultaneous localization and mapping (SLAM) algorithm was used for the trunk regions to construct a tree row model using the depth map.

**3.3 Point Cloud**

*3.3.1 Phenotyping.*

Polo et al. proposed a tree segmentation method to extract various properties of apple, pear, and grape trees [Polo 2009]. The LMS-200 LiDAR sensor was used to acquire the point cloud data of a tree row. Several traits such as volume and surface area were extracted using a numerical algorithm. The analysis showed that the experimental values were well correlated with manually measured values. Rosell et al. proposed a 3D reconstruction method of a tree row [Rosell 2009]. The datasets were constructed for various species of apples, pears, grapes, and citrus fruits using the LMS-200 LiDAR sensor. The data measured at different positions were integrated into a 3D map that represented a tree row. The data collected from left and right sides were registered and merged in a semi-automatic manner. Mendez et al. proposed a tree segmentation method for modeling and maintaining the evolution of apple, pear, and grape orchards [Mendes 2014]. Starting from the bottom point, a set of points were fitted to a cylinder that represented the trunk. Using the kd-tree algorithm accelerated the determination of closest points. Considering the fact that the successor branches were thinner than their parents, a new branch was generated, and a new cylinder was fitted as the successor branch. The process for balancing the branch clusters refined the hierarchical structure of the whole tree. Das et al. built a sensor suite having a laser range scanner, a multi-spectral camera, thermal imaging, and a GPS enabling hand-carrying and UAV-mounting [Das 2015]. The device was used to monitor four tree properties, morphology, canopy volume, leaf area index, and fruit count, in apple and grape orchards. However, the authors did not describe how to reconstruct the 3D tree row from the video or how to extract those values. Peng et al. proposed an apple orchard mapping system [Peng 2016]. They used a pair of coupled sensors of an RGB camera and LiDAR. The feature points extracted from the RGB image and point cloud were associated using a super-pixel matching technique. The registration across frames was accomplished using the singular value decomposition (SVD) method. Finally, by identifying the location and two diameters of each apple, semantic mapping was achieved. Zhang et al. proposed a branch segmentation algorithm for apple-tree images [Zhang 2020d]. Its aim was to measure the number of branches and extract the branch topology and length. Point cloud data were acquired with backpack LiDAR (LiBackpack DG50). Based on the formal model of TreeQSM (https://github.com/InverseTampere/TreeQSM), the point cloud data were analyzed and segmented into a hierarchical tree topology of first, second, and third-order branches. To mitigate the noise effects of point cloud data, filtering and down-sampling algorithms were applied.

Wahabzada et al. proposed a tree segmentation method to extract various geometric shape properties from grapevine images [Wahabzada 2015]. A 3D laser scanning device was used to obtain the point cloud of a tree. A histogram was calculated, and k-means clustering was applied based on the histogram. Each cluster was classified into berry and rachis regions depending on geometric features. The stem and leaf regions were classified using SVM. Scholer et al. proposed a method for reconstructing grape bunches in 3D and deriving useful phenotypic traits [Scholer 2015]. Using a laser scanner of Perceptron ScanWorks v5, they acquired the point cloud for a grape bunch consisting of berry, pedicel, and peduncle regions in the BBCH89 period, during which the berries were fully ripe for harvest. Since the stem was invisible in the season, scanning was performed twice, first scanning the complete grape cluster and then removing the berries and scanning the stem systems. The berries regarded as a sphere were detected using RANSAC. Then, the rachis and other components were extracted using heuristic rules. The reconstruction quality was enhanced using relational growth grammar. Finally, the phenotypic descriptors were extracted. The same research group extended the work to start the processing at an earlier stage, BBCH73, with groat-sized berries and mostly visible stems [Mack 2017].

Zhu et al. proposed a tomato canopy reconstruction method using multi-views [Zhu 2023a]. Using three Kinect v2 sensors placed 120° apart, three RGB-D images were acquired. The images were registered to construct a point cloud using the intrinsic shape signature-iterative closest point (ISS-ICP) algorithm. Then contours of the tomato canopy were extracted using the AlphaShape algorithm, and several parts such as the leaf and stem were separated. The processing was performed in three different seasons, flowering, florescence, and fruiting, to monitor the change in phenotypic traits such as tree height, canopy width, and leafstalk angle. The same research group proposed an extended phenotyping system [Zhu 2023b]. In this case, further segmentation of tree organs such as the stalk and leaf was performed, and finer traits related to the tree canopy were calculated.

Cheein et al. proposed a pear tree segmentation method for canopy volume estimation [Cheein 2015]. Four computational geometry methods, namely the convex hull technique, the segmented convex hull technique, cylinder-based modeling, and the 3D occupancy grid technique, were attempted. Experiments compared the pros and cons of four methods with synthetic data and actual pear tree data acquired with LiDAR. The experiment showed that the 3D occupancy grid method had the best accuracy, while the cylinder-based modeling was the fastest.

The system proposed in [Nielsen 2012a] had several sensors, namely LiDAR sensors, a stereo camera, an odometer, a GPS, and a tilt sensor. A peach tree row was scanned using LiDAR and a stereo camera mounted on a vehicle. The authors used a strategy to improve accuracy by fusing the obtained sensor data. To remove the ground points, the RANSAC algorithm was applied. Three basis vectors obtained by PCA were regarded as the orchard row direction, crossing direction, and pointing up direction. From the GPS data, vehicle trajectory was estimated using the Kalman filter. Finally, GMM was used to segment individual trees that were fitted with cylinders.

Underwood et al. proposed an approach for almond tree segmentation and 3D reconstruction of a tree row with the aim of documenting the traits of trees and later retrieving and updating them for tree management tasks [Underwood 2015]. The LiDAR data of a tree row was sliced and processed sequentially by a hidden semi-Markov model (HSMM). This model segmented the individual trees by transitioning among four states: tree, boundary, gap, and small tree. From the segmented individual trees, a height signature descriptor was extracted and stored in the tree database. Through the recognition model, trees were later retrieved and updated to maintain the information system of an orchard. The same research group proposed an extended version, with the input being the top-view LiDAR point cloud covering an entire apple farm [Bargoti 2015]. Based on the vehicle heading angle, the whole farm was divided into tree rows, and then each tree was processed using HSMM. Tree trunks were detected by applying the Hough transform. This research group further improved the system by adding pixel density features of flowers and fruits [Underwood 2016]. These features were extracted from color images by a simple rule set. Additionally, the same research group proposed a graph-based processing algorithm [Westling 2021a]. The point cloud data of avocado and mango trees were captured using a handheld GeoSLAM Zebedee 1 LiDAR sensor. Synthetic data generated from the SimTreeLS tool [Westling 2021d] was also utilized. Initially, the point cloud was converted into a voxelized data structure and then into a graph. A trunk node was identified, and the remaining nodes found their shortest paths to this trunk node. Each node was scored based on its importance scores. Paths reaching the same trunk node were aggregated and segmented as a single individual tree. The dataset is publicly available and presented in Table A.1 in Appendix.

*3.3.2 Harvesting*

Yongting et al. proposed an apple tree segmentation method for constructing a picking robot [Yongting 2017]. An image was captured with Kinect v2 and transformed into point cloud data using Kinect software. The segmentation algorithm employed both color and point cloud information. It calculated a feature vector comprising the color difference and the distance between the seed point and its neighboring point. A region-growing rule was applied to the seed point to expand a region. This process was iterated until convergence. Subsequently, the feature sets of RGB, HIS, and fast point feature histogram (FPFH) from the point cloud were extracted. These features were classified into fruit, branch, and leaf categories using an SVM, whose hyper-parameters were optimized using the genetic algorithm.

*3.3.3 Spraying*

Mahmud et al. proposed a tree segmentation algorithm to estimate the apple tree canopy foliage density and volume [Mahmud 2021]. A sequence of point cloud data was acquired using a VLP-16 LiDAR sensor with a scanning speed of 5 frames per second. In preprocessing, points corresponding to the ground were removed using a RANSAC-based outlier detection algorithm. The trellis wires were quickly identified using a nearest neighbor search algorithm based on a kd-tree. The support poles and tree trunk were recognized based on their vertical and cylindrical characteristics. The tree area was subdivided into four vertical sections, and the number of points in each section was counted, excluding the trellis wire, pole, and trunk. These counts were used to estimate the canopy density, which in turn was utilized to control the spraying amount of pesticide.

*3.3.4 Pruning*

Karkee utilized point cloud data obtained with a time-of-flight (TOF) camera (CamCube 3.0) to segment an apple tree image [Karkee 2014]. By leveraging the distance quality and amplitude information encoded in each pixel, background pixels were removed. The resultant 3D points were skeletonized using the medial axis algorithm. The interconnections among the skeletons were analyzed, and a tree structure was constructed. By examining the pruning rules of expert humans, a set of rules was developed to determine the pruning points on the skeletal tree structure. Elfiky et al. employed Kinect v2 to capture the front and back sides of an apple tree and generated point cloud data using Kinect Fusion software [Elfiky 2015]. To merge two point clouds into one, the authors proposed an algorithm that combined skeletonization with the iterative closest point (ICP) algorithms. This method of obtaining point cloud data was significantly less expensive than using LiDAR. A stacked-circle algorithm was applied to fit a 3D tree model of the trunk and branches to the point cloud and finally determined the branch growing direction to decide the cut points. Zeng et al. proposed an apple tree segmentation method aimed at constructing a pruning or spraying machine [Zeng 2020]. An image was captured with a VLP-16 LiDAR sensor from a distance of approximately 3 m, containing approximately 4500 to 8300 points per image frame. From this point cloud data, the background was removed using the RANSAC algorithm. Based on the fact that the trellis wires formed a plane, a simple rule was established and applied to detect the trellis. Points outside the trellis plane were considered part of the tree canopy. The trunk and support poles were identified based on their cylindrical shape.

You et al. proposed a cherry tree segmentation method for constructing a pruning robot operating during the dormant season [You 2022a]. This method was tailored specifically for upright fruiting offshoot (UFO)-shaped trees. A custom stereo camera was employed to capture the point cloud. The point cloud underwent skeletonization, and points were classified into five categories: trunk, support branch, leader branch, side branch, and none, utilizing both topological and geometric constraints. To eliminate false branches, the authors used a simple convolutional neural network that processed a $16 \times 32$ image to determine validity. A simple pruning heuristic was implemented to sever all side branches emanating from the leaders. The same research group expanded upon this work by proposing improved decision-making for cutting points and conducting a field test [You 2022b]. They reported a cutting success rate of 58%.

Li et al. proposed a 3D reconstruction method for jujube dormant trees [Li 2023a]. Three RGB cameras captured images that were utilized by the structure from motion (SfM) algorithm to reconstruct the dense point cloud. The SIFT descriptor was employed for feature matching. A significant limitation was the necessity of placing a white curtain behind the tree to increase the success rate of SIFT feature matching.

Westling et al. proposed a segmentation method for mango and avocado trees for pruning tasks [Westling 2021b]. They applied a processing stage similar to that used in their previous study [Westling 2021a], which was aimed at handling phenotyping tasks. The previous study utilized a graph to identify the shortest paths from nodes to the trunk node and aggregated them to identify individual trees. To switch to the pruning task, they estimated the shade score at each node and used it to simulate the pruning effect. The higher the shade score, the more likely the node was identified as a cut point. This simulation involved considering the extent to which opening up the canopy would allow light to pass through the tree after pruning.

*3.3.5 Yield Estimation*

Zine-El-Abidine et al. proposed an individual apple tree segmentation method for counting the apples on the tree [Zine-El-Abidine 2021]. They utilized two seasonal point clouds, one captured at harvest time and the other during the winter. Since the winter image revealed the entire branches, it facilitated the segmentation of individual trees in orchards. Using the harvest image, apples on the tree were detected based on thresholding color channels. They proposed a method that mapped the fruits onto the individual tree. A major limitation was that images captured in different seasons had to be paired and manually registered.

Aiming to circumvent the use of expensive LiDAR sensors, Dey et al. employed the SfM technique to recover the color point cloud from an uncalibrated sequence of grapevine images [Dey 2012]. The reconstruction was accomplished using the bundle adjustment algorithm and multi-view stereopsis. Salient features were extracted from the point cloud using both color and shape information. Using SVM, each point was classified into three categories: berry, branch, and leaf. To reduce noise, a conditional random field (CRF) was applied.

*3.3.6 Navigation*

Zhang et al. proposed a method for extracting tree row information and detecting trunks within the row to guide a navigational robot [Zhang 2013]. A sequence of tree row images was acquired using a custom-built LiDAR sensor. The point cloud was registered and refined based on odometry measurements using the iterative closest point (ICP) algorithm. Using RANSAC, two straight lines were fitted to the left and right sides of the tree row and further refined with an extended Kalman filter. The tree trunks were detected using the particle filter algorithm, which incorporated both the motion data from odometry and the point cloud information.

*3.3.7 Thinning*

Nielsen et al. proposed a method for reconstructing peach trees aimed at facilitating blossom thinning [Nielsen 2012b]. Three RGB cameras were utilized to capture the images. From these images, the point cloud of the tree was recovered using the trinocular stereo technique. The images were captured at night using high-intensity illumination, and the color information was mapped onto the point cloud to produce a color point cloud. However, the study did not describe the process for segmenting the images.

**3.4 Others**

*3.4.1 Phenotyping.*

Chene et al. proposed a depth image segmentation algorithm for apple and yucca trees [Chene 2012]. In a well-controlled indoor situation where each tree was isolated from other objects, they used Kinect to obliquely capture an RGB-D image of the tree. Only the depth map was used for segmentation. The maximally stable extremal region (MSER) algorithm was used to segment the depth map into individual leaf and branch regions. Because of the indoor constraints, the technique was not applicable to an orchard field test.

*3.4.2 Harvesting.*

With the aim of improving the success rate of robot picking, Jin et al. proposed a method to accurately localize the grape ears and cutting point of a grape stem [Jin 2022]. They captured a far depth map using a binocular CCD camera and a close depth map using a RealSense camera installed above the robot hand. The color images from the binocular camera were analyzed and thresholded to identify the grape ear. The robot hand moved close to the grape stem and RealSense captured a close image. The depth map was thresholded to isolate the grape ear and stem. By analyzing the edge segments, the stem was identified and a picking point was selected.

Qiang et al. proposed a citrus tree segmentation algorithm to safely guide a picking robot in a complex natural environment [Qiang 2011]. A multi-spectral, five-channel image of a citrus tree was acquired using a custom-built camera. The minimum noise fraction (MNF) transform was applied to convert this five-channel image into an $n$-channel image using PCA. In their experiments, four-channel images were obtained. A reference spectrum was made by collecting ROIs corresponding to branch regions. A spectral angle mapper was used to convert the four channels into a single spectral angular value, α, by calculating the similarity between the image spectra and reference spectra. Branch regions were then identified by thresholding the α map.

Tanigaki et al. proposed a cherry tree segmentation method that used infrared and red laser sensors [Tanigaki 2008]. To make the problem easier to solve, they assumed "single trunk training," which results in a tall tree with fruits placed around the trunk. They used an infrared 830 nm laser beam to measure the distances to tree parts and a red 690 nm laser beam to detect the red fruit. The red laser was not reflected well by the unripe fruit, leaves, and stalks, but was reflected well by the red ripe fruit. Based on this, a threshold technique was used to segment the red fruit, leaves, and trunk.

Bac et al. proposed a pepper plant segmentation method that used multi-spectral images for the obstacle avoidance of a picking robot [Bac 2013]. An image was labeled with five classes: the stem, leaf top, leaf bottom, fruit, and petiole. The background was removed by thresholding the near-infrared wavelength. For each pixel, a 46-dimensional feature vector was extracted and input to a CART classifier. They attempted to optimize the segmentation accuracy by applying a feature selection technique. The same research group improved the system by enhancing the stem localization capability [Bac 2014].

Colmenero-Martinez et al. proposed a trunk detection method for olive tree images to select the shaking point for a shake-and-catch harvesting machine [Colmenero-Martinez 2018]. An infrared LED scanner mounted on the end-effector was used to capture the image. A sequence of images was analyzed in real-time to decide where to guide the end-effector and when to stop and start shaking.

*3.4.3 Pruning.*

Chattopadhyay et al. proposed an apple tree segmentation method that used multiple depth images acquired with a TOF sensor [Chattopadhyay 2016]. From an apple tree, multiple depth images were acquired at slightly different positions. Each depth image was thresholded to remove the background, and the results were cleaned using a noise removal technique. Using the clean depth image, the system extracted 2D skeletons and estimated the diameter and center of a cross-section of each skeletal pixel to obtain a 3D point cloud. All of the obtained point cloud data were aligned and combined to obtain a 3D tree model. Their dataset is publicly available and introduced in Table A.1. The same research group presented another approach that used the same dataset [Akbar 2016]. The new approach used only one depth image to reconstruct the 3D tree model. The branch diameter was also estimated. The same research group presented another 3D tree modeling algorithm that used 40 depth images per tree acquired with an LMS 111 camera [Medeiros 2017]. A split-and-merge clustering algorithm divided the points into three classes: trunk, junction, and branch points. The trunk and branch points were modeled with cylindrical shapes. Considering the importance of the diameters of the trunk and branches in planning the pruning, these values were estimated using the cylindrical models.

Botterill et al. developed a tree growing grammar set and used it to parse the grapevine image captured by a trinocular stereo camera [Botterill 2013]. To ease the edge detection of each RGB image, the vines were placed on a blue background. The cane edge segments, along with their thickness information, were detected using edge detection. Then, grammatical rules representing the tree structure were applied that recursively joined adjacent cane segments and joined the segment pairs into cane parts. The cane parts were recursively joined to form the whole tree structure. SVM was used to decide whether two segments should be joined or not. Three tree structures were combined into a 3D tree model using a bundle adjustment technique. The same research group extended their work into a complete system by including the processes of deciding the cut points and planning the obstacle-avoiding robot trajectory [Botterill 2017]. Luo et al. proposed a segmentation algorithm for the peduncles of grape clusters using depth images [Luo 2016]. These depth images were captured by a binocular stereo camera. The peduncle was localized as a good candidate for a cutting point. The centers of grape berries were also identified. Then, the 3D spatial coordinates of the cutting points and berry centers were determined using the correspondence between the left and right images. Finally, by aggregating this information, the bounding volume of the grape cluster was calculated and used for selecting the 3D cutting point.

## 4 DL ALGORITHMS FOR TREE IMAGE SEGMENTATION

This section has the same structure as Section 3. The only difference is that this section reviews the studies that used the DL approach. The studies reviewed in this section used the DL models explained in Sections A.1.3, along with the model learning method.

Table 2 summarizes the 82 DL-based tree image segmentation papers. Sections 4.1–4.4 strictly follow the structure of Table 2. The point cloud method was the most commonly used method in studies based on the RB approach. However, in the DL approach, only a few studies used a point cloud. This was because with DL, RGB or RGB-D images are sufficient for obtaining a practical performance. A point cloud is much more expensive to acquire, and DL models using the point cloud method are not better than models using RGB images.

Table 2: Summary of DL-based tree image segmentation papers

| Image Type | Agricultural Tasks and Relevant Papers |
|---|---|
| RGB | **Phenotyping**: [Gene-Mola 2020a](apple), [Sun 2022](apple), [Suo 2022](apple), [Zhao 2023](apple), [Dias 2018a](apple, peach, pear), [Siddique 2022](apple, peach, pear), [Xiong 2023](citrus)<br>**Harvesting**: [Kang 2019](apple), [Kang 2020](apple), [Wan 2023](apple), [Jiang 2023](apple), [Kok 2023](apple), [Kalampokas 2020](grape), [Kalampokas 2021](grape), [Wang 2023a](grape), [Wu 2023a](grape), [Kim 2022](tomato), [Rong 2022](tomato), [Rong 2023](tomato), [Kim 2023](tomato), [Liang 2020](litchi), [Chen 2021a](litchi, passion fruit, citrus, guava, jujube), [Zhong 2021](litchi), [Peng 2023](litchi), [Lin 2022](guava), [Wang 2023b](Sichuan pepper), [Zheng 2023](jujube), [Williams 2019](kiwi), [Williams 2020](kiwi), [Song 2021](kiwi), [Zheng 2021](mango), [Fu 2022](banana), [Wan 2022](pomegranate), [Li 2022](longan)<br>**Spraying**: [Kim 2020](pear), [Seol 2022](pear)<br>**Pruning**: [Tong 2022](apple), [Tong 2023](apple), [Williams 2023](grape), [Gentilhomme 2023](grape), [Borrenpohl 2023](cherry)<br>**Yield estimation**: [Hani 2020](apple), [Gao 2022](apple), [Wu 2023b](apple), [La 2023](apple), [Palacios 2022](grape)<br>**Navigation**: [Wen 2023](pear, peach)<br>**Thinning**: [Hussain 2023](apple), [Majeed 2020b](grape), [Majeed 2020c](grape), [Wu 2022](banana) |
| RGB-D | **Phenotyping**: [Dong 2020](apple), [Milella 2019](grape), [Digumarti 2019](cherry), [Barth 2018](Capsicum annuum (pepper)), [Barth 2019](Capsicum annuum (pepper)), [Barth 2020](Capsicum annuum(pepper)) |

| | **Harvesting**: [Zhang 2018](apple), [Zhang 2019](apple), [Zhang 2020a](apple), [Zhang 2020c](apple), [Granland 2022](apple), [Coll-Ribes 2023](grape), [Xu 2022](cherry tomato), [Zhang 2022b](tomato), [Zhang 2023c](tomato), [Li 2020](litchi), [Yang 2020](citrus), [Lin 2019](guava), [Lin 2021a](guava), [Lin 2021b](guava), [Yu 2022](pomegranate) <br> **Spraying**: [Cong 2022](citrus) <br> **Pruning**: [Chen 2021b](apple) <br> **Training**: [Majeed 2020a](apple) |
|---|---|
| Point cloud | **Phenotyping**: [Dong 2023](apple), [Schunck 2021](tomato, maze) <br> **Harvesting**: [Luo 2022](grape) <br> **Pruning**: [Ma 2021](jujube) |
| Others | **Phenotyping**: [Uryasheva 2022](apple), [Liu 2023](tomato) <br> **Yield estimation**: [Hung 2013](almond) |

### 4.1 RGB

Because of the high accuracy of DL segmentation models based on RGB images, many researchers have used RGB images. Figure 5(b) and Table 2 demonstrate this fact. Some studies used RGB-D images as inputs but used a depth map for thresholding with the aim of removing the farthest trees. In this case, because core segmentation models use RGB images, we introduce them in this section. For example, the papers of [Kang 2019, Lin 2022, Kim 2020] belong in this category.

*4.1.1 Phenotyping.*

Gene-Mola et al. proposed a combination method that performed fruit segmentation using DL models and the 3D reconstruction of apple tree rows using SfM [Gene-Mola 2020a]. A mask R-CNN was employed for the fruit segmentation. For the 3D reconstruction of a tree row, several tens of images were acquired and used by SfM. Another module that projected the detected apples onto the 3D tree row model was developed. Their image dataset is publicly available and shown in Table A.1. Sun et al. proposed a trunk segmentation method for apple trees with the purpose of estimating the diameter of a grafted apple tree trunk [Sun 2022]. An RGB image was labeled with the trunk region, where each trunk had a ground truth for the diameter measured 10 cm above the grafting position. SOLOv2 was employed as a segmentation model. Suo et al. proposed a seedling segmentation method with the aim of grading apple seedlings [Suo 2022]. An RGB image was acquired by placing an apple seedling on white paper. The image was labeled with four classes for the root, rootstock, graft union, and scion. The BlendMask model was employed to segment the seedling image, while seedling quality grading was also performed. Zhao et al. proposed an apple tree segmentation method for estimating five phenotypes [Zhao 2023]. Instead of using expensive LiDAR or RGB-D sensors, they used the heterogeneous cameras installed in a smartphone. An RGB image was labeled with boxes for the fruit, graft, trunk, and whole tree. Additionally, five phenotypes for the trunk diameter, ground diameter, tree height, fruit vertical diameter, and fruit horizontal diameter were manually measured and labeled. YOLOv5s was used to detect the boxes. Based on the detection results and binocular vision, five phenotypes were estimated. Dias et al. proposed a flower segmentation method for apple, peach, and pear tree images with the purpose of estimating the bloom intensity [Dias 2018a]. An RGB image was labeled with the flower region. An FCN was employed for the segmentation. A unique feature of this study was the pursuit of a versatile model by training it with apple tree images and then testing the same model with peach and pear tree images. Experiments showed the good generalizability of the model. The dataset for the apple tree images is publicly available and introduced in Table A.1. The same research group proposed a flower segmentation method for apple, peach, and pear tree images for the purpose of yield prediction based on the flower densities [Siddique 2022]. Considering the scarcity of manually labeled data, they proposed a self-supervised learning (SSL) technique based on contrastive learning. The importance of SSL in the agricultural domain will be discussed in Section 5.

Xiong et al. proposed a citrus tree segmentation method [Xiong 2023]. An RGB image was labeled with three classes for the fruit, branches/leaves, and background. BiSeNetv1 was employed to segment the regions. The segmentation results supported a visual-inertial SLAM-based VINS-RGB-D module, which obtained a semantic point cloud and finally produced a 3D tree row map.

*4.1.2 Harvesting.*

Kang et al. used a Kinect sensor to acquire RGB-D images, but the apple picking robot they constructed only used RGB images [Kang 2019]. They accomplished fruit detection using a modified gated feature pyramid network (GFPN). After the fruit detection, DasNet-v1 was used to segment the fruit and branch regions. A total of 800 images were collected and labeled. The same research group extended their work by adopting DasNet-v2 with ResNet-101 as a backbone [Kang 2020]. The model was trained to perform instance segmentation for fruit and semantic segmentation for stems. After the segmentation, the system constructed a 3D map that was used for determining the fruit pose and 3D working space of the robot. Wan et al. proposed a method for segmenting the branches in apple tree images with the aim of more accurate path planning for a picking robot [Wan 2023]. To enhance the robustness of their dataset, the images were collected on sunny and cloudy days under various illumination conditions. An RGB image was labeled with two classes: branches and other. An ESP and U-net combination was used to construct a new DL model called U$^2$ESPNet to improve the segmentation accuracy. Jiang et al. proposed a method for the segmentation and reconstruction of the thin wires located behind branches and leaves with the aim of avoiding these wires in the fruiting wall architecture of an apple orchard [Jiang 2023]. An RGB image was labeled with the wire regions. The BlendMask model was employed to segment the image. Wire pixels were stitched and fitted using a polynomial function to extract accurate skeletons. Kok et al. proposed an apple branch segmentation method with the aim of avoiding collisions with a picking robot [Kok 2023]. An RGB image was labeled with branch and background regions. U-net++ was employed to segment the branch regions. Using an additional depth image and geometrical constraints, 3D branch surfaces were also reconstructed.

Kalampokas et al. proposed a grapevine segmentation method with the aim of building a grape picking robot [Kalampokas 2020]. An RGB image was labeled with three classes for the grapes, leaves, and background. Eleven different CNN models were trained, and their performances were compared. It was shown that MobileNetv2 produced the best results. Their image dataset is publicly available and shown in Table A.1. The same research group extended their study to include stem segmentation [Kalampokas 2021]. They showed that a regression CNN (RegCNN) model improved the accuracy. Wang et al. proposed a grape fruit and nearby branch segmentation method [Wang 2023a]. An RGB image was labeled with three fruit varieties, including Muscat Hamburg, Chardonnay, and Summer Black. Additionally, it was labeled with three region classes for peduncles, branches, and leaves. The authors designed a new DualSeg network that was composed of a local-processing CNN and global-processing transformer modules. A feature fusion module was also presented. Wu et al. proposed a stem detection method for grapevine images [Wu 2023a]. A grape bunch was labeled with a box and stem was labeled with three key points. YOLOv5 was used to detect the grape bunch and the HRNet was used to identify the stem key points.

With the aim of guiding a tomato picking robot without damaging other fruits or stems, Kim et al. proposed a 6D pose estimation method for tomato plants [Kim 2022]. They combined an EfficientPose network and state-of-the-art 6D pose model into the Deep-ToMatoS network, which performed multiple tasks that included estimating the tomato maturity and identifying the 6D pose of the tomato and stem. A robot control method based on the detected information was also introduced. Rong et al. proposed a method for tomato bunch detection and grasping pose estimation to improve the picking success rate [Rong 2022]. YOLOv5m was employed to detect the tomato bunches and identify the maturity of an individual tomato. A picking sequence planning strategy based on the grasping pose information was also presented. Rong et al. proposed a tomato picking point detection method [Rong 2023]. An RGB image was labeled with three classes for the fruit, calyx, and stem. Swin transformer v2 with an UperNet decoder was employed to segment the regions. The segmented regions were post-processed with thinning and morphological operations to select the picking point. With the aim of determining the easiest grasping poses for picking tomatoes, Kim et al. proposed a method to identify the poses of all the tomato and pedicel pairs in a scene [Kim 2023]. An RGB image was labeled with four key points representing the fruit center, calyx, abscission, and branch. An OpenPose network was used as a backbone to detect the key points and estimate the multiple-tomato poses.

Liang et al. proposed a fruit bunch and stem segmentation method for litchi tree images to select picking points [Liang 2020]. They adopted YOLOv3 to detect a fruit cluster area with a bounding box. U-net was used to segment and crop a 150 × 150 ROI centered in a bounding box. The system used a night images with controlled illumination. With 645 images, they compared nine combinations of three illumination levels (high, normal, low) and three distances (0.6 m, 0.8 m, 1 m) and found the optimal combination. Chen et al. proposed a 3D tree row reconstruction method for a variety of fruits [Chen 2021a]. Experiments were performed with litchi, guava, passion fruit, and citrus tree images. A unique feature of the method was the integration of stereo vision and SLAM techniques to reconstruct a 3D model of a tree row. First, an EfficientDet network was used to detect the fruit regions in RGB images. These detected fruits regions were matched using a dynamic stereo vision technique, and the results were converted into point cloud data. Finally, the global trajectories were estimated by processing the local point clouds extracted from each frame, and they were stitched using an ORB-SLAM3 framework to produce a global 3D map. Zhong et al. proposed a fruit and fruit-bearing branch segmentation method for the purpose of building a litchi picking robot [Zhong 2021]. An RGB image was labeled with fruit and fruit-bearing branch regions. YOLACT was trained using the labeled dataset. A module that determined the picking point and roll angle of a robot was also presented. Peng et al. proposed an accurate segmentation method for litchi branches with the aim of enabling clamping and shearing by a litchi picking robot [Peng 2023]. To improve the segmentation accuracy, they developed a novel DL model called ResDense-focal-DeepLabv3+, which enhanced DeepLabv3+ by incorporating a ResNet and DenseNet with focal loss. The dataset was divided into three levels of complexity (simple, medium, and complex) to evaluate the network.

The guava picking system developed by Lin et al. used RealSense D435i, but only used the RGB channels [Lin 2022]. The authors labeled 891 guava tree images with three classes for the fruit, branches, and background. They appropriately modified MobileNet to optimize it for the segmentation of these three classes and embedded it with channel- and spatial-attention modules. The feature aggregation process was also applied to detect thin branches.

Wang et al. proposed a Sichuan pepper fruit detection and branch segmentation method [Wang 2023b]. An RGB image was labeled with a box for fruit and regions for the fruit and branches. A neural network inspired by YOLOP was developed to simultaneously solve the multiple tasks of detection and segmentation.

Zheng et al. proposed a trunk and branch segmentation method to identify an appropriate shaking point for a jujube tree [Zheng 2023]. An RGB image was labeled with three classes for the trunk, branches, and background. AGHRNet, which improved the HRNet by embedding a ghost attention module, was proposed to segment the regions. The best shaking point was identified using the depth information.

Williams et al. proposed a kiwi picking robot [Williams 2019]. The robot had four arms working simultaneously. It was an improvement of the system described in [Scarfe 2009]. The robot worked in a kiwi orchard with a pergola style frame. The authors fine-tuned the pre-trained FCN to segment three classes for the calyx, cane, and wire. The calyx identification was advantageous in stereo matching and selecting gripping points. The paper also introduced a robot arm scheduling algorithm for kiwi cluster picking. The same research group improved the robot system by using faster R-CNN and better end-effector [Williams 2020]. The improved system achieved a picking success rate of 86% and 2.78 s/fruit, while the previous system provided 51.0% and 5.5 s/fruit. Song et al. proposed a kiwi picking robot with a camera 1 m below the tree canopy looking upward [Song 2021]. Using DeepLabv3+ with a ResNet-101 backbone, the image was segmented into four classes for the calyx, branches, wires, and background. To tackle the imbalance among these four classes, uniform and median-frequency weights for cross entropy losses were compared. It was concluded that uniform weights were better. After segmenting, a trellis was detected using a progressive probabilistic Hough transform (PPHT).

Zheng et al. proposed a fruit segmentation and picking point detection method for mango harvesting [Zheng 2021]. An end-to-end DL model was proposed that simultaneously segmented the fruit region and detected the best picking point. A mask R-CNN was modified such that it had two additional outputs for the fruit region and picking point.

Fu et al. proposed a bunch and stalk detection method for banana tree images [Fu 2022]. An RGB image was labeled with a pair of boxes for the bunch and stalk. YOLOv4 was used to detect the boxes. The images were collected under various weather conditions such as sunny, cloudy, and overcast.

Wan et al. proposed a pomegranate tree segmentation method for a picking robot [Wan 2022]. A total of 1000 artificial pomegranate tree images were synthesized and 200 actual tree images were collected with RealSense D435. YOLOv4 with a CSPNDarknet53 backbone was employed to detect branches using bounding boxes. The boxes were merged using distance and angle constraints to identify a single long branch region. Finally, polynomial fitting was applied to the branch region.

Li et al. proposed a fruit-bearing branch segmentation method for longan tree images for selecting the picking points [Li 2022]. An RGB-D image was acquired with RealSense D455 carried by a drone. The RGB image was input to the YOLOv5s model to detect the box that included the fruit and fruit-bearing branch. The box was further segmented into a branch region using DeepLabv3+. Based on the branch region and depth map, the pose and picking point were selected.

*4.1.3 Spraying.*

Kim et al. developed a spraying system for a pear orchard [Kim 2020]. The system captured RGB-D images with RealSense D435. They input the RGB images into SegNet and segmented them into five classes, including leaf-branch-trunk, fruit, ground, sky, and pipe regions. Using the depth map, the pixels for distances greater than 2 m were regarded as the trees in another row and eliminated. The system worked at 15 fps. The same research group extended their work to have three spraying modes: all open, on/off, and a variable flow rate [Seol 2022]. Their experiments showed that the three modes reduced the pesticide used to 56.8%, 39.37%, and 8.08% in areas where no tree was placed, respectively.

*4.1.4 Pruning.*

Tong et al. proposed a trunk and branch segmentation method for the robotic pruning of apple trees [Tong 2022]. A total of 2000 images were collected in the dormant season after the leaves had fallen off. An RGB image was labeled with three classes for the trunk, branches, and support. Two models that used a cascade mask R-CNN with ResNet-50 and Swin-T as a backbone were trained and evaluated. The branch and trunk regions in a binary map were skeletonized using a Zhang-Suen thinning algorithm. The junctions between the trunk and branches were identified as pruning points. The same research group proposed an improved method that used SOLOv2 [Tong 2023]. This model was superior to the mask R-CNN by 19% in terms of the mAP. Furthermore, their paper presented a pruning point decision method that used a rule set that worked with the depth information obtained by a RealSense D435i camera.

With the aim of building a cane pruning system, Williams et al. proposed a grapevine segmentation method [Williams 2023]. An RGB image was labeled with four classes for the trunk, cane, wire, and node. The image was segmented with a Detectron2 network. Using two segmented images of the left and right sides, a 3D tree model was constructed and used by a decision maker for cane pruning. Gentilhomme et al. proposed a grapevine branch segmentation method to assist in accurately understanding the vine structure and planning pruning points [Gentilhomme 2023]. An RGB image was captured with a smartphone camera after placing a white curtain as the background. This RGB image was labeled with five classes, including those for the trunk, courson, cane, shoot, and lateral shoot. ViNet, which was an improved version of a stacked hourglass network (SHN), was proposed to identify the node positions and their relationships. A graph was constructed with the node information. A shortest path algorithm was applied to perform the branch region segmentation. Their dataset is publicly available and described in Table A.1.

Borrenpohl and Karkee proposed a method to segment the trunk and leader regions in a cherry tree image [Borrenpohl 2023]. In an orchard with an upright fruiting offshoot (UFO) tree architecture, two RGB tree images were captured with active and natural lighting during the dormant season. An RGB image was labeled with three classes for the trunk, leader, and background. A mask R-CNN was used to segment the regions. A method to estimate the leader diameter was also presented. Their experiments showed that the active lighting images produced better accuracy.

*4.1.5 Yield Estimation.*

Hani et al. proposed an apple tracking method for apple counting in an apple tree row [Hani 2020]. The work was an extension of [Roy 2018]. While [Roy 2018] used GMM, [Hani 2020] employed both GMM and U-net and compared their performances. They captured a video along a tree row from the front and back sides. Each video was processed by U-net to segment the apples. By applying a tracking algorithm, the detected apples were tracked along a sequence of frames, and more accurate counting was accomplished. They reported that their DL method produced better accuracy in apple counting. As a byproduct, a 3D reconstruction of the apple tree row was produced. Gao et al. proposed a fruit and trunk detection and tracking method for accurately counting apples [Gao 2022]. An RGB image was labeled with boxes for the apples and trunk. The YOLOv4-tiny model was used to detect them. Unlike other detection-and-tracking approaches that tracked the fruits, this method tracked the trunk because the trunk was much larger and easier to track. For the fruit tracking, the reference displacement between consecutive frames was estimated and used to predict the possible fruit locations. The same research group improved their work by proposing a fruit tracking technique with mutual matching [Wu 2023b]. La et al. considered an orchard where adjacent trees were heavily intertwined and proposed a tree region segmentation method for apple tree images [La 2023]. An RGB image was labeled with a tree region by drawing the contour of an individual tree. YOLOv8 was employed to segment this tree region. The dataset is publicly available and described in Table A.1.

A berry detection and canopy segmentation approach was proposed with the aim of counting the actual number of berries in a grape cluster [Palacios 2022]. An RGB image was acquired and labeled with six classes: gap, leaf abaxial, leaf adaxial, shoot, trunk, and cluster. Using SegNet, a series of segmentation operations for the clusters, individual berries, and canopy elements were performed. Considering hidden berries, the number of detected berries was regressed to obtain the actual number of berries using a support vector regressor (SVR).

*4.1.6 Navigation.*

Wen et al. proposed a segmentation method for pear and peach orchard images [Wen 2023]. An RGB image was labeled with several classes, including road, person, fence, wall, ground, ladder, and fruit tree classes. Object segmentation outside the tree regions made robot navigation a potential application task. The authors proposed a transformer-based model called MsFF-SegFormer to segment the regions. For each object region, depth information was estimated using a depth map.

*4.1.7 Thinning.*

Hussain et al. proposed a green apple segmentation and pose estimation method with the aim of thinning the green apples [Hussain 2023]. An RGB image of green apples in the size of 10 to 30mm was captured with a smartphone. The image was labeled with green fruit and stem regions. A mask R-CNN was used to segment the regions. The orientations of the fruit and stem were estimated by applying PCA to each fruit and stem region. The pose information for the connected fruit and stem was aligned and refined using Harris corners.

Majeed et al. proposed a grapevine segmentation algorithm with the aim of selecting an end-effector pose to automate the green shoot thinning task [Majeed 2020b]. Images were captured around the first week of the bud-opening season before the leaves appeared. Thus, the cordons were very visible. A total of 191 images were labeled using three classes for the trunk, cordon, and background. A pre-trained SegNet was transfer-learned to the tree-image dataset. Because the background class accounted for approximately 98%, a median-frequency class balancing technique was applied. The segmented map was further processed to divide the left and right cordons with respect to the trunk. Each cordon was skeletonized and fitted with the $6^{th}$ polynomial. The same research group developed an extended system that dealt with grapevine images captured between the first week of bud opening and the fourth week when the cordon was only partially visible because of the foliage [Majeed 2020c]. They used a faster R-CNN to detect the trunk and cordon bounding boxes. The $6^{th}$ polynomial was fitted based on the centroids of the cordon bounding boxes.

With the aim of cutting off the male flower cluster of a banana tree, Wu et al. proposed a method for localizing the cut point [Wu 2022]. An RGB image was labeled with rectangular boxes for the bananas, male flower clusters, rachis, and stem. YOLOv5s with an improved loss function was used to localize them. A cutting point was selected based on the depth information.

## 4.2 RGB-D

*4.2.1 Phenotyping.*

Dong et al. proposed a 3D reconstruction method for an apple tree row and phenotyping method for measuring tree traits such as the tree volume, canopy volume, trunk diameter, and apple count [Dong 2020]. Videos were captured with an Intel RealSense R200. Each of the front and back side RGB-D videos was processed to construct a 3D map, and the two resulting maps were merged into one. The SIFT features and RANSAC were used to process each video. A mask R-CNN was used to segment the trunk. The trunk information was used to support the merging of the two 3D maps and measure the trunk diameter.

Milella et al. proposed a grapevine segmentation method for canopy volume estimation and bunch counting [Milella 2019]. Two RGB-D images were acquired using RealSense R200 devices mounted on a moving vehicle, one mounted laterally and the other in a forward looking direction. These two RGB-D images were transformed into a point cloud. The point cloud was used to reconstruct a 3D tree row. Additionally, an RGB image was labeled with five classes: bunch, pole, trunk-cordon-cane, leaf, and background. VGG19 was employed to classify the patches into one of these five classes. Various traits of individual trees were measured by combining the 3D map and classification results.

A segmentation simulation for synthetic tree images was introduced in [Digumarti 2019], whose results could be applied generally. The images for six tree species, including cherry and five non-fruiting trees, were synthesized using the SpeedTree software for tree modeling and the Unreal Engine for rendering. A total of 4800 images per species were generated. Each image was labeled automatically using six classes for the trunk, branches, twigs, leaves, ground, and sky. SegNet was used with three different inputs, including RGB, an early fusion of RGB and depth, and a later fusion of feature maps from the RGB and HHA networks. The HHA network processed an HHA image constructed from a depth map by extracting the horizontal disparity, height above ground, and angle. Their experiments showed that the later fusion gave the best accuracy.

A research team at Wageningen University and Research (WUR) published a series of papers proposing dataset synthesis for capsicum annum plants [Barth 2018]. An image was labeled automatically with eight classes, including the background, leaf, pepper, peduncle, stem, shoot and leaf stem, wire, and cut peduncle regions. For comparison purposes, 50 actual images were captured and labeled manually. The same research group presented a segmentation algorithm for a synthetic dataset [Barth 2019]. A practical training strategy was proposed in which a large synthetic dataset was used for bootstrapping a CNN model and a small actual dataset was used to fine-tune the model. They published another paper that improved the synthetic dataset using a generative DL model like cycle GAN [Barth 2020].

*4.2.2 Harvesting.*

Zhang et al. proposed an apple tree segmentation algorithm with the aim of selecting the shaking points for a shake-and-catch system [Zhang 2018]. Images were acquired of a fruiting wall architecture using Kinect v2 during the dormant season. An image was labeled with two classes for the branches and background. The depth image was converted into a pseudo-color image using a jet color model. The region maps from the pseudo-color image and depth image were combined to improve the accuracy. The regions were segmented using an R-CNN with AlexNet as the backbone. The experiment showed that combining the images improved the accuracy. Because dormant season images are not appropriate for harvesting purposes and the input for AlexNet was small $32 \times 32$ images, the same research group proposed an improved version for a shake-and-catch harvesting machine [Zhang 2019]. A total of 253 apple tree images were newly acquired using Kinect v2. An image was labeled with three classes for the trunk/branches, fruit, and leaves. A pre-trained Deeplabv3+ with ResNet-18 as a backbone was employed and fine-tuned. The same research group published another paper that extended the scope when selecting the shaking point [Zhang 2020a]. In the new research, 785 new apple tree images were acquired using Kinect v2. An image was labeled with four classes for the branches, trunk, fruit, and background (mostly leaves). The Deeplabv3+ with ResNet-18 backbone was used to segment the regions. For each of the trunk and branch regions, a curve was fitted. Their intersection was identified as a shaking point, and the mean thicknesses of the trunk and branches were used to refine the shaking point. The same research group presented a different approach that used a faster R-CNN as the object (fruit, branch, and trunk) detection model [Zhang 2020c]. The boxes for the branches and trunk were fitted with skeleton segments. The shaking point was then determined based on these segments. Granland et al. proposed a semi-supervised learning method for apple tree segmentation to accomplish multiple tasks, including harvesting, thinning, and pruning [Granland 2022]. Because manual labeling is very expensive, a semi-supervised method that uses a small number of labeled samples and large number of unlabeled samples is needed.

One way to accomplish this is to use human-in-the-loop labeling, where a model is initially trained with a small number of labeled samples, and then unlabeled samples are segmented and a human corrects the results. The paper also proposed an automated repair algorithm called automating-the-loop. As a segmentation model, it adopted U-net with a ResNet-34 backbone, which took a four-channel RGB-D image as the input. The repair process consisted of filtering, tree fitting, and repair. Experiments showed that the automating-the-loop method was competitive with the human-in-the-loop method.

Coll-Ribes et al. proposed a bunch and peduncle detection method for grapevine images [Coll-Ribes 2023]. An RGB image was labeled with three classes for the bunches, peduncles, and background. The RGB and depth channels were concatenated and input to a mask R-CNN to segment the regions. The diameter of a bunch was measured and the cut point for a peduncle was selected based on the depth information.

Xu et al. proposed a fruit and stem segmentation method for selecting the picking point for a cherry tomato plant [Xu 2022]. An RGB image was labeled with fruit bunch pair, stem, and background regions. An improved mask R-CNN was employed. It used a combination of RGB and depth channels as the input and had two output heads for the fruit and stem. Zhang et al. proposed a 3D pose estimation method for a tomato bunch with the aim of guiding a picking robot [Zhang 2022b]. The main stem, fruit stem, and fruit were modeled using a cylinder, two connected cylinders, and sphere, respectively. A box and 11 key points were labeled per tomato bunch to localize them. An Hourglass model was employed to detect them. The detected information was finally transformed into 3D pose information, which was used to guide the picking robot to cut the bunch. The same research group improved their work by proposing a detection method for invisible key points [Zhang 2023c].

Li et al. proposed a litchi tree segmentation method for a picking robot [Li 2020]. A total of 452 images were collected with Kinect v2. An image was labeled with three classes for the twigs, fruit, and background. The twig regions were post-processed with morphological operations and skeletonization. After selecting the fruit-bearing twigs, 3D information was recovered using a point cloud. Finally, the medial axis was estimated using PCA, which was used to select the picking point.

Yang et al. proposed a fruit detection and branch segmentation method for citrus tree images [Yang 2020]. Images were collected under different lighting conditions, including front, side, and back lighting. Each image was labeled with the fruit and branch regions. A branch was labeled with two types of labeling, overall and segmental. A mask R-CNN model was used to segment the regions. The whole branch and trunk regions were identified by merging branch segments. The fruit and branch diameters were also estimated based on the depth information.

Lin et al. proposed a guava tree image segmentation and fruit pose estimation method for robot picking [Lin 2019]. A total of 437 RGB-D images were collected by Kinect v2. An image was labeled with three classes for the fruit, branches, and background. First, an image was segmented into fruit and branch regions by fine-tuning an FCN with a VGG-16 backbone. Using the depth map, the fruit region was further separated into individual fruits, and a branch region was converted into a 3D line segment. Finally, by analyzing the geometric relationship between an individual fruit and mother branch, the fruit pose was estimated. The pose information was used for planning a collision-free robot path. The same research group improved the system by modifying a series of processing stages [Lin 2021a]. They collected 304 new guava tree images. An image was labeled with three classes as in the previous paper. Because the dataset was small, an improved DL model, a tiny mask R-CNN that used a tiny CNN as a backbone with eight convolution layers and five pooling layers was employed. The fruit and branch regions were converted into a point cloud. The fruit regions were fitted with spherical shapes using RANSAC. Unlike the previous work that reconstructed a 3D line segment from a branch point cloud, the new system constructed a cylindrical model using PCA. A 3D tree model was built using the spherical fruit and cylindrical branches. In the successive paper, they also proposed a robot path planning algorithm that used the 3D fruit and branch information [Lin 2021b]. Because the future robot path was heavily dependent on the past path, the authors regarded the path planning problem as a Markov decision process. They adopted a reinforcement learning model and trained the model with a deep deterministic policy gradient (DDPG), coupled with long short-term memory (LSTM). A simulation and field test were performed to evaluate the success rate of robot picking. In the field test, the grasp, detach, and harvest success rates were 90.16%, 68.85%, and 59.02%, respectively.

Yu proposed a pomegranate fruit detection method [Yu 2022]. A mask R-CNN was used to segment an RGB-D image into fruit regions. The region information was converted into a point cloud, which was processed with PointNet. Finally, 3D boxes representing fruits and the surroundings such as branches and leaves were obtained.

*4.2.3 Spraying.*

A citrus tree crown segmentation algorithm was proposed for the purpose of spraying [Cong 2022]. A special aspect of this study was the use of images taken during different seasons that included the seeding, flourishing, and fruiting stages. A total of 766 images were acquired with Intel RealSense d435i. A mask R-CNN with an SE attention embedded ResNet as a backbone was developed. Each image was preprocessed to remove the regions at the back row using a thresholding depth map. The system worked at 5 fps to accomplish real-time processing for the spraying task.

*4.2.4 Pruning.*

Chen et al. proposed an apple tree branch segmentation algorithm that used RGB-D images acquired with an Intel RealSense D435 [Chen 2021b]. The results could be applied to the branch pruning and fruit thinning tasks. They labeled 512 images with two classes for branches and non-branches. Four-channel RGB-D images were input into neural networks. Three models, Pix2Pix based on a generative adversarial network, U-net, and DeepLabv3, were compared and the superiority of U-net was shown. In the case of highly occluded branches, Pix2Pix was better. To measure the performance under the occlusion, they presented a new performance metric called the occlusion difficulty index that used a third class called "occluded branch."

*4.2.5 Training.*

Trunk and branch segmentation is critical to accomplish automatic tree training. Majeed et al. proposed an apple tree segmentation algorithm with the aim of automating the tree training task [Majeed 2020a]. The SegNet segmented the image into three classes for the trunk, branches, and trellis. They preprocessed an RGB-D image to obtain a foreground RGB image by removing pixels whose depth was greater than 1.3 m. By comparing the results when using the original RGB and foreground RGB images, they concluded that using the foreground RGB image was superior.

**4.3 Point cloud**

*4.3.1 Phenotyping.*

Dong et al. proposed a method for phenotyping the 3D traits of individual apples [Dong 2023]. A UAV with three cameras obliquely captured images of an apple tree row. The images were captured using three different tree training systems. A dense point cloud was constructed using the multi-view structure obtained from a motion algorithm based on bundle adjustment. A modified U-net that used voxels as an input produced feature maps in the embedding space. A clustering algorithm was applied in the embedding feature space to segment individual apples. The apples were fitted with spheres. The apple count, volumes of individual apples, and spatial density distribution map were estimated as phenotypic traits.

Schunk et al. constructed a dataset to study the growth processes of tomatoes and maze and used it to develop a base DL model for tree segmentation [Schunck 2021]. They periodically captured point clouds for each growing stage and labeled them with two classes for leaves and stems. Tree segmentation was performed using PointNet. Then, spatio-temporal registration and surface reconstruction were attempted, and the base performance was reported.

*4.3.2 Harvesting.*

With the aim of preventing a picking robot from colliding with other objects, Luo et al. proposed a method for grapevine segmentation and 6D pose estimation [Luo 2022]. An RGB camera and LiDAR were used to capture grapevine images. From the RGB image, grape clusters were segmented using a mask R-CNN. A point cloud was then segmented based on the RGB segmentation results. Finally, the pose of a grape cluster was estimated by peduncle surface fitting, and the best cutting point was located on the peduncle ROI.

*4.3.3 Pruning.*

Ma et al. proposed a jujube tree segmentation method for identifying a branch to be pruned, along with the best cutting point [Ma 2021]. A jujube tree image was acquired with two Azure Kinect DK cameras mounted 35 cm apart and synchronized on a common supporting frame. First, the nearby trees and background were removed by thresholding the distance values and post-processing to obtain a clean point cloud. Another image was captured at the opposite side to supplement the first image in order to obtain a 3D reconstruction. The reconstruction was accomplished by coarse registration based on skeleton points. The iterative closest point (ICP) algorithm was applied for fine registration. The 3D tree was further segmented into individual branches to select the cutting points. An SPGNet with four steps that included geometric partitioning, super-point graph construction, super-point embedding, and contextual segmentation was adopted to achieve the trunk and branch segmentation.

**4.4 Others**

*4.4.1 Phenotyping.*

Uryasheva et al. proposed a leaf segmentation method with the aim of monitoring the health of apple trees [Uryasheva 2022]. They used three cameras to obtain multi-spectral images of apple trees during the blooming season when the trees were most susceptible to fungal infection. To increase the robustness of the system, the images were acquired under multiple environmental conditions. An image was labeled with two classes for the leaves and background. An Eff-Unet, which combined an EfficientNet and U-net, was employed to segment the leaf regions of one-channel images. The segmentation maps from multiple channels were registered and used to calculate various vegetation indices (VI) such as NDVI, NDRE, and GNDVI.

To support multiple tasks like picking, pruning, and pollinating, Liu et al. proposed a stem segmentation method for tomato plant images [Liu 2023]. An RGB and NIR camera pair was used to capture images. The YOLACTFusion model, which integrated feature maps from the RGB and NIR images, was developed to enhance the feature discrimination ability. Experiments showed that fusing RGB and NIR images produced a superior segmentation accuracy.

*4.4.2 Yield Estimation.*

Hung et al. proposed an almond tree segmentation method that used multi-spectral RGB–NIR images [Hung 2013]. An image was labeled with five classes for the fruit, trunk, leaves, ground, and sky. Feature learning was performed in an unsupervised manner using a sparse autoencoder neural network. Instead of classifying each pixel, the correlation of neighbor pixels was considered, with a conditional random file (CRF) used for classifying pixels. A comparison of the RGB only and RGB–NIR images showed that the additional information provided by the NIR data enhanced the segmentation accuracy.

**5 DISCUSSION AND FUTURE WORK**

Based on the reviews in Sections 3–4, this section discusses the research trends and future perspectives regarding three aspects: sensors, methods, and datasets. Finally, important future research directions are given.

## 5.1 Sensors

As shown in Figure 5(b), the modern DL approach generally uses RGB or RGB-D images. This is primarily because a DL model guarantees a high segmentation accuracy when using RGB or RGB-D images. Actually, many studies captured RGB images using a smartphone. Because a farmer already has a smartphone, this would be very advantageous, with no extra cost for the sensor. For some tasks like picking, pruning, and thinning, an RGB-D sensor is essential because a robot needs 3D information for travelling to the target position. The most common sensor for RGB-D is an Intel RealSense D435, which costs approximately 200 USD. A LiDAR sensor for a point cloud and the sensors for multi-spectral or hyper-spectral images are very expensive, while no clear merit in relation to the segmentation accuracy is gained. Therefore, in the future, it is expected that more researchers will use RGB or RGB-D images. One exception is when a very accurate phenotype is required because a LiDAR sensor provides more accurate 3D information. The theme of monocular depth estimation, which infers the depth from a single RGB image will be presented as a future work.

## 5.2 Methods

As explained in Figure 5(a), the major solutions for tree segmentation are rapidly transforming from the RB to DL methodology. The DL model is changing from a CNN to transformer. Because a transformer has a less inductive bias than a CNN, the transformer is better for building a more versatile system. The fusion of a CNN and transformer will be presented in a future work.

The current tree segmentation methods are very specific to a given task and environment. Therefore, a new task in a new environment requires a high cost for designing, training, and testing a new method. This fact acts as a great barrier to broadly applying computer vision to tree segmentation. In the AI community, this kind of barrier is overcome by developing a uniform model with multi-modality capability. One notable example is Uni-perceiver, which can process text, image, and video data for classification, detection, and segmentation tasks using a single unified model [Zhu 2022]. In the agricultural domain, very limited attempts have been made in this direction. To the best of our knowledge, fruit or flower segmentation for several different species using a single model is the only example [Siddique 2022]. The development of a versatile model in the agriculture domain will be presented in a future work.

The 3D reconstruction of a row of trees is very useful for globally monitoring and managing orchard information. Combining it with individual tree segmentation will be beneficial for many agricultural tasks. The reconstruction method still relies on RB such as SfM. The pursuit of DL-based reconstruction will be presented as a future work.

## 5.3 Datasets

A labeled dataset is an essential ingredient for building a successful DL model. In the medical imaging research communities, there have been many notable datasets, some acting as de facto standards [Li 2023b]. In agriculture, there have surveys of datasets [Lu 2020, Luo 2023]. However, the list is deficient in quantity and quality. Because of the diversity resulting from numerous factors such as the fruit type, orchard environment, training type, time (day or night), season (dormant or foliage), image type, and viewing angle, the existing datasets listed in Table A.1 of Section A.3 are limited in the sense that they only cover specific tasks and environments. The construction of versatile datasets will be presented as a future work.

Prominent recent themes in artificial intelligence are few-shot learning and self-supervised learning [Gui 2023]. These learning strategies are very effective in overcoming the lack of human-labeled data because only a limited amount of labeled data is needed to enable a massive amount of labeling to be performed automatically. The employment of these learning methods for tree segmentation will be presented as a future work.

## 5.4 Future works

Based on the three aspects mentioned above, six future works are summarized as follows.

- Monocular depth estimation infers a depth map from a single RGB image or video. For natural images with applications in augmented reality or autonomous driving, many excellent methods are available [Ming 2021]. However, few papers can be found in the agriculture domain. Cui et al. produced a vineyard depth map from an RGB video using U-net, which is a good basis to begin research in the agriculture domain [Cui 2022]. The resulting maps could be used for tasks such as robot path planning and visual servoing.
- Fusing a CNN and transformer has been proved to extract more robust features [Khan 2023]. Medical imaging is actively adopting fused models to identify thin vessels and small tissues [Chen 2021c]. Agricultural tasks could benefit from a fused model to improve the segmentation accuracy, especially in identifying small fruits, thin branches, and highly occluded fruits or flowers.
- If a versatile model is developed for agriculture, it would be applicable to various tasks, either immediately or after light fine-tuning. Specific constraints could be exploited when developing a unified model for tree segmentation, including the fact that the objects in the image are confined to tree components such as the fruit, flowers, branches, leaves, and trunk. These constraints will make the development much easier than that for a random scene.
- The 3D reconstruction of a row of trees could be implemented by combining DL techniques. For example, combining panoramic imaging and monocular depth estimation from an RGB video is expected to accurately generate a 3D tree row model.
- Table A.1 in Section A.3 presents 11 public datasets related to tree image segmentation. Except for the last dataset, each one is highly specific to a given task. The construction of versatile datasets with large quantities and good quality is a critical element in future tree segmentation research. These datasets can serve as de facto standards to enable the objective comparison of newly developed segmentation models. Additionally, they will motivate challenging contests.
- In the agriculture domain, this is just an initial stage of adopting few-shot learning or self-supervised learning [Yang 2022, Siddique 2022, Guldenring 2021]. These strategies will be very effective in overcoming the obstacles incurred as a result of the inherent large variations in agricultural tasks and environments. They will promote a more versatile system. It is highly recommended that a pre-trained model with a large and general agricultural dataset be built and then fine-tuned to handle downstream segmentation problems suitable for given tasks using few-shot learning.

## 6 CONCLUSIONS

Using a crawling review, we collected 158 papers on fruit tree image segmentation for various agricultural tasks. At the top level of the taxonomy, the papers were grouped by their use of RB (76 papers) and DL (82 papers) methods. They were further classified systematically according to image, task, and fruit species. We hope that readers grasped the big picture of the broad spectrum of tree segmentation research. The discussion and future works focused on overcoming the current bottleneck imposed by the lack of a versatile dataset and segmentation model.

Fruit tree segmentation is closely related to agricultural tasks, requiring a certain level of expertise in agriculture. Therefore, to build successful automation machines, the close collaboration of computer scientists and horticulturists is very important. We hope this review will motivate each of these communities and lead to close collaboration. This interdisciplinary work will pave the way to build a versatile tree segmentation module and lead to the use of numerous high-accuracy agricultural machines in orchards.

35

# A APPENDICES

In the appendix section, three topics are described.

## A.1 Basics of Image Segmentation and Performance Metrics

The localization of interesting objects in images is an essential problem for computer vision. This problem is solved in two different ways, detection and segmentation. A detection algorithm identifies an object using a rectangle, while segmentation designates a set of pixels covering an object. This paper focuses on segmentation.

This section briefly introduces the most popular image segmentation algorithms. These algorithms are divided into two methodologies, traditional RB and modern DL methods. In an RB method, humans manually design the detailed procedure of a segmentation algorithm using their own reasoning and heuristics. In contrast, a DL method trains a deep neural network using data. Because of its flexibility and high performance, DL became dominant in the early 2010s. Although RB methods now have diminishing value, we include them to present a complete review.

In general, an RB method is not *semantic* in the sense that it only outputs regions without object class information. In contrast, a DL method is semantic because it produces object class information with confidence values. Tree segmentation can be regarded as a special case of image segmentation because only one object class, "tree," is considered. Therefore, even the RB approach is semantic because an image is segmented into foreground and background regions, and the foreground is regarded as the tree object. Often, the foreground region is then further segmented into several parts such as the fruit, branches, trunk, and leaves.

In DL, general image segmentation aims at identifying numerous classes. For example, the YOLACT model can identify 80 classes of objects such as a person, car, bicycle, and cat. Two different types of segmentation are supported, semantic and instance segmentation. They are the same in the sense that object regions are segmented and each region is labeled with an object class. However, when multiple instances of the same class are found in an image, semantic segmentation treats them as a whole by giving them the same identifier, while instance segmentation differentiates them by giving them different identifiers [Kirillov 2019]. Depending on the requirements of an agricultural task, we may choose between semantic and instance segmentation. For example, in building a spraying robot aiming at reducing the amount of pesticide used by targeting only the tree regions, semantic segmentation is chosen. In building a pruning robot, instance segmentation is chosen because the target of the pruning is an individual tree.

### A.1.1 Image Type

Because of improvements in image sensing technologies, a variety of image sensors are available and various image types are used for tree segmentation. We classify these into RGB, RGB-D, point cloud, and others. An RGB image consists of three channels with red, green, and blue wavelengths. An RGB-D image has an additional channel representing the depth. A point cloud image only stores the depths at points where the sensor can measure the distance of light travel. Usually, a point cloud is too expensive to acquire but provides more accurate depth values than RGB-D. Multi-spectral and hyper-spectral images belong to the "others" image type. A multi-spectral image requires more than three wavelengths, usually resulting in tens of channels. A hyper-spectral image has hundreds or thousands of channels. Sonar, thermal, and depth images also belong to this type.

RGB, RGB-D, multi-spectral, and hyper-spectral images are represented using a grid-like structure (i.e., a 3D array) because every pixel in the 2D image space has a value. In contrast, a point cloud image is represented by a 3-tuple set $(x,y,d)$, where $(x,y)$ is the image position and $d$ is its depth at that position. Some sensors such as LiDAR cannot measure depths at some positions as a result of the physical limitations of laser light and use a point cloud.

Because an RGB camera is very cheap, with every farmer already possessing a high-resolution camera installed in their smartphone, and numerous high-performance deep learning models pre-trained with RGB images are publicly available, RGB images are the most popular. Since the first release of Kinect in 2010, many vendors have provided cheap and high-resolution RGB-D cameras at a cost of several hundred dollars. Kinect from Microsoft and RealSense from Intel are popularly adopted for agricultural applications. Because the LiDAR camera needed to produce a point cloud and a multi/hyper-spectral camera are very expensive, their usage is limited to special application environments. In the RB era, because segmentation algorithms were far from the performance level needed for practicality, some researchers directed their attention to point cloud or multi/hyper-spectral images, which provide richer information. However, in the DL era, this attention has dramatically diminished and RGB and RGB-D images have become dominant. Currently, numerous deep learning models pre-trained with RGB or RGB-D images are available, and transfer learning to a tree image dataset is easily accomplished. Figure 5(b) and Table 2 show this trend. A specific survey paper focusing on RGB-D for fruit localization is available [Fu 2020].

*A.1.2 Rule-based Methods*

From the beginning of the computer vision era, researchers have developed many image segmentation algorithms [Szeliski 2022]. The basic strategy is to group nearby pixels with similar intensity, color, and/or texture features into a region or to divide the image into successive regions by delineating contours where the features abruptly change. Some of the methods that have been popularly used for tree segmentation are briefly explained below.

**Thresholding**: The thresholding algorithm divides the pixels into two groups below and above a threshold. One of the two groups is taken as the foreground (object), in our case the tree region, and the other is considered to be the background. In implementing the thresholding algorithm, the most important decision concerns the optimal threshold value. Intuitively, a valley point in a histogram is a good candidate, but noise prevents this scheme from working successfully. Otsu proposed an optimization method that minimizes the weighted sum of the variances of the pixel groups in the foreground and background regions. Because the thresholding operation is applied to a single map, an appropriate channel for the RGB or RGB-D images must be selected.

**Clustering**: A pixel or super-pixel is represented by t-tuples $(v_1, v_2, \ldots, v_t)$, where $v_i$ is a color or position value. The clustering algorithm groups the pixels or super-pixels into several clusters. Each cluster represents foreground or background region. Often a finer-segmentation is accomplished by assigning each cluster to one of the branch, trunk, leaf, fruit, or background classes. Many clustering algorithms are available, such as k-means, Gaussian mixture model (GMM), fuzzy, and mean-shift.

**Region growing**: Starting from a seed region, the algorithm merges neighboring pixels under some constraints to grow the region. We regard the snake algorithm as belonging to the region growing category because it successively expands the object contour by optimizing the smoothness and objectness of regions.

**Machine learning**: After representing a pixel with a t-tuple, as in the clustering approach, a machine learning model is trained with a human-labeled training set. At the inference stage, the pixels in a new image are classified into several classes such as branch, fruit, leaf, and trunk classes. The Bayesian, support vector machine (SVM), multi-layer perceptron (MLP), and random forest are popularly used learning models. The main difference compared to DL is that the model here has a shallow architecture (i.e., 1–3 layers), while DL models have tens or hundreds of layers. Because the layers are shallow, the feature learning is weak. Therefore, the feature vector is designed by humans heuristically. In contrast, DL models extract the optimal features by learning, resulting in a feature learning capability.

**Fitting**: Assuming geometric shapes such as a lines, circles, and cylinders for the tree parts, a fitting method finds the optimal parameters for those shapes for the given point set. The Hough transform, random sample consensus (RANSAC), and geometric fitting algorithms belong to this approach.

**Graph-based**: A pixel becomes a node of a graph and adjacent nodes are connected by edges whose weights represent the similarity between two pixels. A graph cut algorithm is applied recursively to the graph to partition the nodes into groups. Each group is regarded as a region. Sometimes, a line segment obtained using a skeletonization algorithm becomes a node, and node merging is applied successively to obtain a tree structure.

*A.1.3 Deep Learning Methods*

Figure A.1 illustrates the basic input and output relationship of a DL model used for fruit tree image segmentation. In this case, the output map is represented using three channels for the branches, fruit, and background. Some other region maps are illustrated in Figure 2.

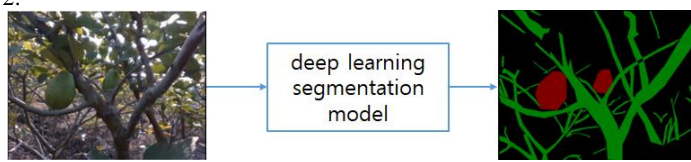

Figure A.1. Outline of DL segmentation (a guava tree image is segmented into three classes: fruit, branches, and background [Lin 2022]).

In 2012, AlexNet achieved a top-5 classification error rate (15.3%) in the ImageNet large scale visual recognition challenge(ILSVRC) and won first place Krizhevsky 2012]. This event motivated most computer vision research groups to change their methodology from RB to DL. After this successful application of a CNN to image classification, the first successful object detection model (R-CNN) [Girshick 2014] was proposed in 2014, and its performance continues to improve [Minaee 2022]. The R-CNN is a two-stage model that first generates numerous region proposals and then selects the true regions from them. Because of its generate-and-select strategy, it is slow. In 2017, the R-CNN was transformed into a segmentation model called a mask R-CNN, which has two heads, one for detection and the other for segmentation [He 2017]. Since its advent, the mask R-CNN has popularly been used for segmentation purposes. In 2016, a one-stage detection model called YOLO was proposed [Redmon 2016]. It is fast because it uses one stage of processing. The performance of YOLO has continued to improve from v1 to v8. Like the mask R-CNN, YOLOv8 is a dual-function model for detection and segmentation. In contrast to the R-CNN and YOLO models, which started as detection models and evolved into segmentation models, a fully convolutional network (FCN) was born in 2015 as a segmentation model [Long 2015]. The FCN was designed to use an original image as an input and output a segmentation map. After the success of the FCN model, many variants such as DeConvNet, U-net, and DeepLabv3+ have been proposed and popularly used. After the introduction of a transformer model for language processing by [Vaswani 2017], the transformer model has successfully evolved to achieve image segmentation.

CNN-based models follow.

**mask R-CNN**: The R-CNN was developed as an object detection model and has evolved into fast R-CNN and faster R-CNN [Ren 2015]. The pipeline of these models consists of a backbone, region proposal, and head. The backbone module extracts a rich feature map. The region proposal module generates several candidate patches with a high probability to be meaningful objects. The head module produces the "class" (object class and confidence information) and "box" (object location) branches. Adding the "mask" branch results in a mask R-CNN, which can simultaneously detect and segment the objects [He 2017].

**YOLOv8:** In contrast to a slow two-stage R-CNN that generates many region proposals and successively evaluates each one, YOLO uses a one-stage process that directly regresses the object locations from each window in a grid-partitioned image [Terven 2023]. Owing to its simpler architecture, YOLO achieves real-time processing while sacrificing accuracy. The original YOLOv1 has been continually improved by the originator and other research groups. The latest version, YOLOv8, expands the detection-specific model into a versatile model that enables detection, segmentation, pose estimation, and tracking [YOLOv8].

**YOLACT**: YOLACT is the same as YOLO in that it uses one-stage processing and guarantees real-time processing. However, it uses a completely different approach with two modules, one generating prototypes and the other computing the mask coefficients [Bolya 2019]. The modules work in parallel, resulting in a high speed. The prototypes are feature maps of the same size as the input image, each of which is likely to represent the appearance of certain objects. Linearly combining the prototypes weighted by the mask coefficients results in the final segmentation map.

**FCN**: The input to an FCN is the original image and the output is a $k$-channel binary map, where each channel represents the regions of an object class. In the three-class segmentation shown in Figure A.1, a three-channel map is used for the fruit, branch, and background regions. The encoder part of an FCN reduces the resolution, and the decoder part restores the original resolution. This restoration is achieved using the transpose convolution.

**U-net**: A U-net is a variant of an FCN, which was originally developed for medical image segmentation [Ronneberger 2015]. It adds a skip connection in order to enrich the feature maps transferred from the encoder to decoder. At present, the U-net is used universally for images generated in many fields, including tree images.

The transformer model is innovative in extracting rich feature maps by measuring the self-attention between words in an input sentence [Vaswani 2017]. It has revolutionized natural language processing. The vision community modified the transformer model to make it suitable for processing images by considering the grid patches to be words and succeeded in obtaining superior performances compared to a CNN in many vision problems like image classification, object detection, segmentation, and tracking [Khan 2022]. Here, we introduce well-known transformer models with segmentation capability.

**DETR**: The detection with transformer (DETR) model extracts a $d$-channel feature map with ResNet [Carion 2020]. Each channel is regarded as a word and input into the encoder block, which computes a self-attention map using a query-key-value operation. After passing several encoder blocks, the feature map is passed to a series of decoder blocks, where the feature map is transformed into object location and class information.

**Swin transformer**: The Swin transformer was developed to serve as a backbone for various tasks such as classification, detection, and segmentation [Liu 2021]. It employs hierarchical multiscale feature maps, which start from small patches and gradually merge with neighboring patches in deeper layers. The algorithm employs the shifted-window concept, allowing the overlapping of window partitions.

Model training and performance metrics are explained in the followings.

**Model training**: A tree segmentation model segments a tree image into 2–6 region classes. For example, La et al. segmented an apple image into two classes, the foreground (tree) and background, as seen in Figure 2(a) [La 2023]. Hussain et al. segmented an apple image into three region classes, the fruit, stem, and background, as seen in Figure 2(d) [Hussain 2023]. It is a common practice for researchers to select an appropriate DL model and modify its head to suit the number of region classes.

In the training phase, two approaches are available: transfer learning and learning from the scratch. Transfer learning fine-tunes a pre-trained deep learning model using a tree segmentation dataset. It keeps the weights of the feature extraction layers of the pre-trained model while initializing the weights of the new head layer. The fine-tuning does not significantly perturb the weights of the existing layers by keeping the learning rate very low, while setting the learning rate for the newly attached head layers to be relatively large. The low learning rate conserves the feature extraction capability of the existing layers. The learning-from-scratch method initializes the weights of all the layers and starts the learning from scratch. A reasonable learning rate is applied over all of the layers.

Although there are several public datasets, as explained in Section A.3, most studies use private datasets. This is because the existing datasets are specific to a given task and particular situation. The construction of a general dataset seems to be the most important future work, and this issue is discussed in Section 5. Because the private datasets are small, data augmentation is very important. Usually, a combination of various geometric transformations such as rotation or flipping and various photometric transformations such as adding noise or an intensity change is applied to augment the data.

**Performance metrics**: Once the model has been trained, a performance evaluation follows. Several metrics are popularly used. The pixel accuracy (PA) is measured based on the ratio of correctly labeled pixels to the total number of pixels. The mean pixel accuracy (MPA) is defined as the average PA over all the classes. The intersection over union (IoU) is defined as the area of intersection between the predicted region and ground truth region divided by the union of the two regions. When we fix a threshold value for the IoU, each region can be regarded as true positive (TP), false positive (FP), or false negative (FN). The precision, recall, and F1 score can be computed based on the numbers of TP, FP, and FN regions. Additionally, the average precision (AP) and mean average precision (mAP) are computed by varying the threshold. It is common for AP@0.5 and AP@0.75 to be provided for thresholds of 0.5 and 0.75, respectively. The AP@0.5:0.95 is also used, which represents the average AP using thresholds from 0.5 to 0.95 in increments of 0.05.

## A.2 Agricultural Tasks Supported by Tree Segmentation

It is important to understand the agricultural tasks that require tree segmentation. Table 1 lists these tasks. Section A.2.1 explains the agricultural environments in which tree images are acquired. Section A.2.2 briefly describes each of the tasks in relation to tree segmentation.

### A.2.1 Agricultural Environments

There are several environmental factors that influence the design and implementation of a tree segmentation algorithm.

**Season**: It is important to select the most appropriate season based on the task. Farmers perform pruning during the dormant season. Therefore, for the pruning task, images are typically acquired during the dormant season when the trees have no leaves. This is advantageous for segmenting the branches and selecting the cutting points. Figure 2(c) shows an example. For the spraying task, it is necessary to process an image taken at full foliage. Figure 2(a) shows an example.

**Natural vs. trained tree**: A modernized orchard trains the trees to have a well-controlled shape. This training is advantageous for the health and growth of fruit because it allows a good air flow and sunlight. Additionally, the simpler and rather flat canopies for the trained trees facilitate automating many tasks such as harvesting, pruning, and thinning. A naturally growing tree has a spherical shape, where the back side and interior of the sphere are typically hidden. This makes the segmentation of the fruit, branches, and leaves difficult. The delineation of neighboring trees is also very difficult when the trees are intertwined.

**Day vs. night**: In the daytime, the illumination varies greatly as the weather or viewpoint changes. When a robot changes its pose, the camera can face the sun or becomes backlit. The shadows also matter. These illumination factors degrade the performance. To cope with this difficulty, some researchers use night vision with artificial illumination. For example, Xiong et al. proposed a vision algorithm for a nocturnal image with LED illumination [Xiong 2018a].

**Growing stage**: A fruit tree passes through a long period from the bud to the mature fruit. In the early stages, thinning is a major task, which removes some of the buds, flowers, or fruit. For example, Hussain et al. proposed a fruit and stem segmentation algorithm for thinning the green fruit of an apple tree, as shown in Figure 2(d) [Hussain 2023]. For the harvesting task, images taken at the mature-fruit stage are used.

*A.2.2 Agricultural Tasks*

Next, the characteristics of each agricultural task are briefly introduced, along with several recent survey papers. These survey papers are valuable in overcoming the limit of our survey and deepening the reader's insight and knowledge.

**Phenotyping**: Tree phenotyping is the task of describing the expression of tree traits [Zhang 2023b]. In their review, Huang et al. grouped various traits into five aspects: water stress, architecture parameters, pigments and nutrients, degree of disease, and biochemical parameters [Huang 2020]. The accurate segmentation of individual trees is sufficient to estimate some traits such as the tree height or canopy volume. For other traits such as leaf pigment, disease, or the branching shape, finer segmentation of the fruit, leaves, and branches is required. The academic and industrial activities of the International Plant Phenotyping Network (IPPN) are noteworthy, including the Forest Phenotyping Workshop.

**Harvesting**: Fruit harvesting can be divided into two approaches: bulk harvesting using the shake-and-catch method and selective harvesting by robot picking. The most recent survey papers discuss these [Zhang 2020b, Droukas 2023, Mail 2023, Yang 2023a]. In robot picking, the most important task is fruit detection. Finer segmentation is required to successfully access the fruit without colliding with other fruit, branches, or poles. In the shake-and-catch method, the segmentation of the trunk or main branch is sufficient.

**Spraying**: The main purpose of autonomous spraying machines is to reduce the amount of pesticide by precisely spraying the target tree. The nozzle of the spray gun is only open when it is moving across the tree region [Meshram 2022, Dange 2023]. Because spraying is usually performed by navigating a tree row, real-time tree segmentation is required.

**Pruning**: According to [Zahid 2021], the harvesting and pruning done during apple production account for 59% and 20% of the labor cost, respectively. Therefore, the automation of the pruning task is very important. In this task, the precise identification of branches is very important to analyze the branching structure and select cutting points [He 2018, Zahid 2021, Zeng 2022]. The branch structure is analyzed, and the heuristics that humans use are applied to the branch structure to select the cutting points.

**Yield estimation**: Direct yield estimation is the task of estimating the total amount of fruit on a tree or in a tree row by counting the visible fruits [Koirala 2019, Maheswari 2021, Farjon 2023]. Tree segmentation is essential to perform fruit counting on individual trees. A video is used for fruit counting in a row of trees [Villacres 2023]. The tree regions should be precisely segmented in order to exclude the fallen fruit and fruit in other rows.

**Navigation**: Tree segmentation is required to plan a precise path for vehicle navigation along a tree row in an orchard [Li 2021, Wang 2022]. For tall trees like palms, trunk segmentation is performed. For small trees like apple trees, whole tree regions should be segmented. It is common for information from multiple sensors such as GPS and odometry sensors to be combined with visual information to optimize the path planning.

**Thinning**: Thinning refers to removing some of the blossoms, flowers, or fruit to maximize the quality of the harvested fruit [Lei 2023]. A traditional non-selective method such as a rotating string thinner carries the risk of damaging the tree. Often the thinning results are unsatisfactory because the same operation is enforced over the whole tree without evaluating the quality of each fruit or flower. A vision-based thinning method is selective because it evaluates each one and removes the bad ones. This method should segment each blossom, flower, or fruit and evaluate them.

**Training**: In modern orchards, it is common for fruit trees to be trained to have a controlled shape, which results in a high yield and good fruit quality. The automation of tree training requires the segmentation of branches, leaves, trellises, and poles. For example, Majeed et al. proposed a segmentation method for trunks, branches, and trellis wires to automate apple tree training [Majeed 2020a].

**A.3 Public Datasets**

This section introduces the public datasets related to tree image segmentation that are readily downloadable from the web. All the datasets except the last row were constructed and used by the authors of the papers reviewed in Section 4. Table A.1 summarizes them. Four and three of the 11 datasets are for apple trees and grape vines, respectively. One dataset each was found for tomato plants, avocado trees, and capsicum annum plants. All of these except the capsicum annum dataset contain real images. The last dataset, Urban Street Tree, was not constructed in an orchard, but on the street. However, it is included because fruit trees are available, and the quantity and quality of the dataset are good.

*A.3.1 Apple*

The WACL dataset constructed by Purdue University contains depth images captured from one indoor and four orchard trees [Chattopadhyay 2016]. Diameter information is also provided for each branch in a tree image. The diameter was used to determine whether the branch should be pruned and to select the cutting point. Fuji-SfM is an RGB dataset for apple trees constructed by Gene-Mola et al. at the University of Lleida [Gene-Mola 2020b]. The purpose of the dataset was to reconstruct a 3D tree row from several tens of RGB images. The same research group proposed a reconstruction algorithm and evaluated the performance using the dataset [Gene-Mola 2020a]. The NIHHS-JBNU dataset was constructed through the collaboration of the National Institute of Horticultural and Herbal Science and Jeonbuk National University, Korea [La 2023]. Each RGB image was captured with the target tree in the center of the image. The images were captured in an apple orchard where adjacent trees were heavily intertwined. It was labeled with two regions for the trees and background. The dataset is known to be the first that considered intertwined trees. The Fruit Flower dataset was constructed at Marquette University by Dias et al. [Dias 2018a]. Each RGB image of an apple, peach, or pear tree is labeled with the flower region. To assist in this labeling, a Monte Carlo region growing technique was developed and used [Dias 2018b]. It is worth noting that the research group applied innovative ideas for a versatility test and self-supervised learning using their datasets [Dias 2018a, Siddique 2022].

*A.3.2 Grape*

The Grapes-and-Leaves dataset was constructed at the International Hellenic University by Kalampokas et al. [Kalampokas 2020]. Each RGB image is labeled with three classes for the grapes, leaves, and background. The same research group also produced the Stem dataset, which labels the stems of grape clusters [Kalampokas 2021]. The purpose of these two datasets was to build a grape picking robot by developing a deep learning model that selected the picking point and planned a safe path for robot travel. The 3D2cut Single Guyot dataset was constructed at the Idiap Research

Institute, Switzerland, by Gentilhomme et al. [Gentilhomme 2023]. It has 1513 images that are labeled with five classes for the trunk, courson, cane, shoot, and lateral shoot. Additionally, their node, termination, and spatial dependency are labeled. The purpose of the dataset was to build a pruning machine for a grapevine trained in the guyot system.

*A.3.3 Tomato*

The Pheno4D dataset was constructed with the aim of studying the growth processes of tomatoes and maze at the University of Bonn by Schunck et al. [Schunck 2021]. Seven tomato plants were recorded over three weeks, resulting in 140 point clouds. The images are labeled with two classes for leaves and stems. By separately labeling each leaf, instance segmentation for a leaf was possible. Because the images were captured indoors with the plants growing in separate pots, their application is limited.

*A.3.4 Avocado*

The Avocado dataset, which has a point cloud format, was constructed at the University of Sydney by Westling et al. [Westling 2021c]. Three avocado trees were captured in an image and labeled with three classes for the leaves, branches, and ground. The images at three different stages are provided, one before any pruning, a second after limb removal, and a third after limb removal and hedging. The dataset was used to build a phenotyping method [Westling 2021a] and an automatic pruning machine [Westling 2021b]. The research group also constructed a mango dataset, but they only released the avocado dataset.

*A.3.5 Capsicum Annum (pepper)*

The Capsicum Annum Image dataset was constructed at Wageningen University and Research (WUR) by Barth et al. [Barth 2018]. A dataset of 10,500 RGB-D images was synthesized from 42 plant models by varying plant parameters. An image is labeled with eight classes, including the background, leaf, pepper, peduncle, stem, shoot and leaf stem, wire, and cut peduncle classes. Because the renderer knew the class of each pixel, the labeling of the synthetic images was performed automatically. For comparison purposes, 50 actual images were captured and labeled manually. It was reported that approximately 30 min was needed to label one image. The dataset was used by the same research group for phenotyping [Barth 2019]. The synthetic dataset was improved using a cycle GAN [Barth 2020].

*A.3.6 Street Trees*

The Urban Street Tree dataset includes 41,467 images of 50 tree species, including some fruit-bearing trees. The 22,872 images are divided into several groups, and each group is labeled with a combination of trunk, leaf, tree, branch, flower, and fruit. The authors argued that the dataset could be used for benchmarking multiple agricultural tasks.

Table A.1: Public datasets for tree image segmentation

| Dataset | Fruit | Image | URL |
| --- | --- | --- | --- |
| WACL [Chattopadhyay 2016] | apple | depth | https://engineering.purdue.edu/RVL/WACV_Dataset |
| Fuji-SfM [Gene-Mola 2020b] | apple | RGB | http://www.grap.udl.cat/en/publications/datasets.html |
| NIHHS-JBNU [La 2023] | apple | RGB | http://data.mendeley.com/datasets/t7jk2mspcy/1 |
| Fruit Flower [Dias 2018a] | apple, peach, pear | RGB | https://doi.org/10.15482/USDA.ADC/1423466 |

| Grapes-and-Leaves [Kalampokas 2020] | grape | RGB | https://github.com/humain-lab/Grapes-and-Leaves-dataset |
|---|---|---|---|
| Stem [Kalampokas 2021] | grape | RGB | https://github.com/humain-lab/stem-dataset |
| 3D2cut Single Guyot [Gentilhomme 2023] | grape | RGB | https://www.idiap.ch/en/scientific-research/data/3d2cut |
| Pheno4D [Schunck 2021] | tomato and maize | point cloud | https://www.ipb.uni-bonn.de/data/pheno4d/ |
| Avocado [Westling 2021c] | avocado | point cloud | https://data.mendeley.com/datasets/d6k5v2rmyx/1 |
| Capsicum Annum Image (synthetic) [Barth 2018] | Capsicum annum | RGB-D | https://data.4tu.nl/articles/_/12706703/1 |
| Urban Street Tree [Yang 2023b] | Street trees (50 species) | RGB | https://ytt917251944.github.io/dataset_jekyll/ |